\documentclass[conference]{IEEEtran}
\usepackage{times}

\usepackage[numbers]{natbib}
\usepackage{multicol}
\usepackage[bookmarks=true]{hyperref}
\usepackage{graphics} % for pdf, bitmapped graphics files
\usepackage{epsfig} % for postscript graphics files
\usepackage{mathptmx} % assumes new font selection scheme installed
\usepackage{times} % assumes new font selection scheme installed
\usepackage{amsmath} % assumes amsmath package installed
\usepackage{amssymb} % assumes amsmath package installed
\usepackage{xcolor}
\usepackage{graphicx}
\usepackage{booktabs}
\usepackage{adjustbox}
\usepackage{booktabs}
\usepackage{graphicx}
\usepackage{multirow}
\usepackage{listings}
\usepackage{graphicx}
\usepackage{booktabs}
\usepackage{adjustbox}
\usepackage{url}
\usepackage[utf8]{inputenc}
\usepackage{amsfonts}
\usepackage{algorithm}
\usepackage{algpseudocode}
\usepackage{float}
\usepackage{pifont}
\usepackage{wrapfig}
\usepackage{tabularx}       % 自适应宽度表格
\usepackage{booktabs}       % 专业表格线
\usepackage{makecell}       % 单元格内换行
\usepackage{multirow}       % 多行合并
\usepackage{siunitx}        % 专业单位排版
\definecolor{myRed}{RGB}{255,0,0}
\definecolor{myGreen}{RGB}{0,176,80}
\definecolor{myBlue}{RGB}{53, 158, 213}
\definecolor{myYellow}{RGB}{223, 152, 36}

\begin{document}

% paper title
% \title{{\textbf{\textit{FastUMI}}}: A Scalable and Hardware-Independent \underline{U}niversal \underline{M}anipulation \underline{I}nterface}

% \title{{\textbf{\textit{FastUMI}}}: A Scalable and Hardware-Independent \underline{U}niversal \underline{M}anipulation \underline{I}nterface with Dataset}

\title{{\textbf{\textit{FastUMI}}}: A Scalable and Hardware-Independent Universal Manipulation Interface with Dataset}

% You will get a Paper-ID when submitting a pdf file to the conference system

%%%%%%%% option 1 %%%%%%%% 有西工大的版本
% \author{
% Zhaxizhuoma$^{1\dagger}$, Kehui Liu$^{1,2\dagger}$, Chuyue Guan$^{1,3\dagger}$, Zhongjie Jia$^{1,4\dagger}$, Ziniu Wu$^{1,5\dagger}$, Xin Liu$^{1,4\dagger}$\\
% Tianyu Wang$^{1,6*}$, Shuai Liang$^{1,4*}$, Pengan Chen$^{1,7*}$, Pingrui Zhang$^{1,6*}$, Haoming Song$^{1,4}$, Delin Qu$^{1,6}$, \\
% Dong Wang$^{1}$, Zhigang Wang$^{1}$, Nieqing Cao$^{8}$, 
% Yan Ding$^{1\ddagger}$, Bin Zhao$^{1,2\ddagger}$, Xuelong Li$^{1,9}$ \\
% \\
% $^{1}$Shanghai AI Lab, $^{2}$Northwestern Polytechnical University, $^{3}$Stanford University, \\$^{4}$Shanghai Jiao Tong University, $^{5}$University of Bristol, $^{6}$Fudan University, \\$^{7}$The University of Hong Kong， $^{8}$Xi'an Jiaotong-Liverpool University, 
% $^{9}$Institute of AI, China Telecom Corp Ltd \\
% $\dagger$ $*$ Equal Contribution, $\ddagger$Project Leader, 
% Project Website: \url{https://fastumi.com/}
% }

%%%%%%%% option 2 %%%%%%%% 没有西工大的版本
\author{
Zhaxizhuoma$^{1\dagger}$, Kehui Liu$^{1\dagger}$, Chuyue Guan$^{1\dagger}$, Zhongjie Jia$^{1,2\dagger}$, Ziniu Wu$^{1,3\dagger}$, Xin Liu$^{1,2\dagger}$\\
Tianyu Wang$^{1,4*}$, Shuai Liang$^{1,2*}$, Pengan Chen$^{1,5*}$, Pingrui Zhang$^{1,4*}$, Haoming Song$^{1,2}$, Delin Qu$^{1,4}$, \\
Dong Wang$^{1}$, Zhigang Wang$^{1}$, Nieqing Cao$^{6}$, 
Yan Ding$^{1\dagger\ddagger}$, Bin Zhao$^{1\ddagger}$, Xuelong Li$^{1,7}$ \\
\\
$^{1}$Shanghai AI Lab, $^{2}$Shanghai Jiao Tong University, $^{3}$University of Bristol, \\$^{4}$Fudan University, $^{5}$The University of Hong Kong, $^{6}$Xi'an Jiaotong-Liverpool University, 
\\$^{7}$Institute of AI, China Telecom Corp Ltd \\
$\dagger$ $*$ Equal Contribution, $\ddagger$Project Leader, 
Project Website: \url{https://fastumi.com/}
}

\maketitle

\begin{abstract}
Real-world manipulation data involving robotic arms is crucial for developing generalist action policies, yet such data remains scarce since existing data collection methods are hindered by high costs, hardware dependencies, and complex setup requirements. In this work, we introduce \emph{FastUMI}, a substantial redesign of the Universal Manipulation Interface (UMI) system that addresses these challenges by enabling rapid deployment, simplifying hardware–software integration, and delivering robust performance in real-world data acquisition.
Compared with UMI, FastUMI has several advantages: 1) It adopts a decoupled hardware design and incorporates extensive mechanical modifications, removing dependencies on specialized robotic components while preserving consistent observation perspectives.
2) It also refines the algorithmic pipeline by replacing complex Visual-Inertial Odometry (VIO) implementations with an off-the-shelf tracking module, significantly reducing  deployment complexity while maintaining accuracy. 3) FastUMI includes an ecosystem for data collection, verification, and integration with both established and newly developed imitation learning algorithms, accelerating policy learning advancement. Additionally,
we have open-sourced a high-quality dataset of over 10,000 real-world demonstration trajectories spanning 22 everyday tasks, forming one of the most diverse UMI-like datasets to date.
Experimental results confirm that FastUMI facilitates rapid deployment, reduces operational costs and labor demands, and maintains robust performance across diverse manipulation scenarios, thereby advancing scalable data-driven robotic learning.
% The project is available at: \url{https://fastumi.com}
\end{abstract}

% \IEEEpeerreviewmaketitle

\section{Introduction}

\begin{figure*}[t]
    \centering
    \includegraphics[width=0.9\linewidth]{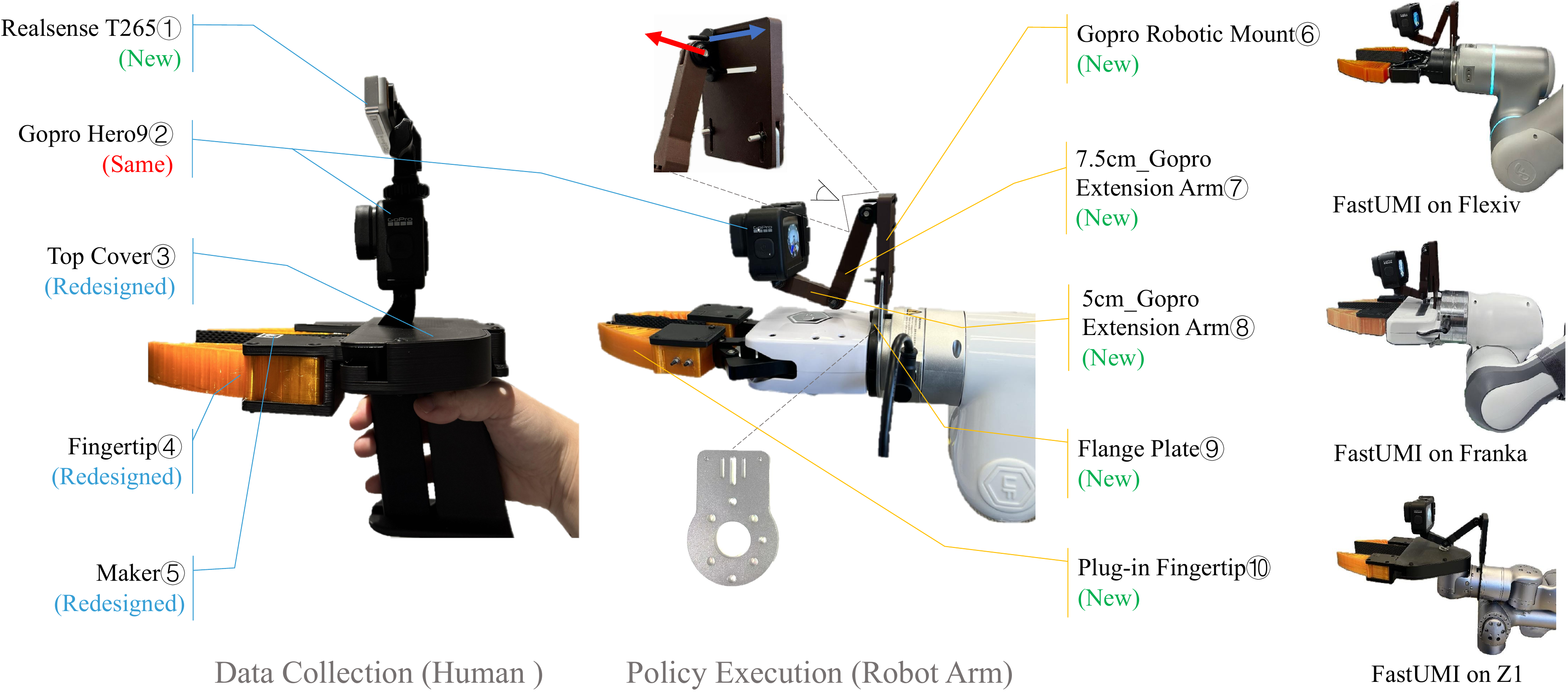}
    \caption{Physical prototypes of FastUMI.
    \textbf{Left}: The handheld device, \emph{used to collect demonstration data from human operators}, includes a GoPro\ding{173} for visual feedback, a RealSense T265\ding{172} for end-effector pose tracking, fingertip markers\ding{175}\ding{176} to measure the gripper aperture, and a top cover\ding{174} to secure both the GoPro and T265.
    \textbf{Middle}: A robot-mounted device, \emph{used for executing learned policies on the robotic arm}, mirrors the handheld configuration.
    It features an ISO-standard-compatible camera mounting solution (including gopro mount\ding{177}, extension arms\ding{178}\ding{179}, and flange plate\ding{180}) that adapts to varying arm and gripper geometries. 
    This design maintains consistent GoPro perspectives across different setups, enabling direct transfer of human demonstration views to autonomous robotic executions. 
    % The robotic arm shown here is an xArm 6.
    \textbf{Right}: FastUMI can be easily deployed on various robotic arms and grippers.
    To distinguish FastUMI's hardware configuration from that of the original UMI, we employ a color-coding scheme.
    % number 11{\scalebox{0.8}{\textcircled{\scalebox{0.7}{\textbf{11}}}}}
    % \textcolor{myGreen}{Green} denotes newly introduced components not present in UMI;
    % \textcolor{myBlue}{Blue} indicates modules redesigned or adapted from UMI's original architecture;
    % \textcolor{myRed}{Red} highlights elements retained from the UMI system.
    }
    \label{fig:overview}
\end{figure*}

The scarcity of large-scale, high-quality, real-world interaction data remains a major bottleneck to progress in robotic manipulation, primarily due to challenges associated with efficient and scalable data collection methods~\cite{bahl2022humantorobotimitationwild,chi2024universal,zhaxizhuoma2024alignbotaligningvlmpoweredcustomized}.
Current methods can be broadly categorized into teleoperation-based techniques~\cite{zhu2023viola,ral_fan_2023,ram_zhang_2024,wu2024gello}, vision-driven demonstrations~\cite{bahl2022humantorobotimitationwild,levine2016learninghandeyecoordinationrobotic}, and sensor-enhanced interfaces~\cite{cabi2020scalingdatadrivenroboticsreward, wang2023mimicplaylonghorizonimitationlearning,zhang2018deepimitationlearningcomplex}. 
While teleoperation enables precise data acquisition, it remains labor-intensive, costly, and constrained by the challenges of non-intuitive and task-specific human-to-robot mapping\cite{zhao2023learning,shi2024yell,fu2024mobile}.
Vision-driven approaches can provide large-scale, low-cost data but typically lack the rich, fine-grained interaction dynamics essential for policy learning.
In contrast, sensor-enhanced interfaces\textemdash exemplified by systems such as the Universal Manipulation Interface (UMI)~\cite{chi2024universal}\textemdash offer a promising alternative. 
They directly capture diverse, multimodal signals that closely align with a robot's onboard sensory modalities, preserving fidelity and precision, while \emph{enabling human demonstration data to be seamlessly transferred into robotic frameworks.}
Moreover, these interfaces can be manufactured and deployed at lower cost, thereby facilitating faster, high-quality data collection under real-world conditions.
In doing so, they narrow the gap between human demonstrations and autonomous robotic execution, enabling rapid and high-quality data collection.

While the UMI system addresses key challenges in human demonstration data collection and supports action policy learning in diverse scenarios, its \emph{current} system design and implementation suffers from two key limitations.
First, its tight coupling with specific robotic components (\emph{e.g.}, the Weiss WSG-50 gripper) restricts adaptability and increases both financial and logistical burdens.
Integrating UMI into different robotic platforms requires not only designated grippers and related hardware, but also extensive efforts, including mechanical redesign, sensor recalibration, and code parameter modification. 
These adjustments impose significant labor overhead and lack generalizability, ultimately hindering widespread adoption across diverse deployment environments and application contexts.

The second limitation arises from the software framework, particularly the reliance on a GoPro-based VIO (Visual-Inertial Odometry) pipeline in conjunction with open-source SLAM algorithms~\cite{campos2021orb}.
Through experimental evaluation, we observe that the UMI system encounters difficulties in tasks that involve prolonged occlusions, such as hinged operations.
As a result, the UMI software configuration struggles to maintain robust operation when visual signals are intermittently lost, thereby diminishing data quality and reducing its utility for subsequent learning tasks. 
% This configuration proves vulnerable when encountering even brief occlusions, and the absence of a high-precision inertial reference further worsens tracking stability.
% As a result, the UMI system struggles to maintain robust operation in scenarios where visual signals are intermittently lost, diminishing data quality and reducing its utility for subsequent learning tasks.
Furthermore, the VIO process is sensitive to parameterization and requires complex calibration procedures and multiple coordinate transformations. 
These factors collectively increase operational complexity, hinder reproducibility, and undermine the user-friendliness of the overall framework.

To address the limitations of UMI, we undertake an extensive redesign centered around three primary objectives:
\begin{itemize}
    \item \emph{Enhancing adaptability through hardware decoupling}.
    By removing strict dependencies on specific robotic components, our hardware design can be seamlessly integrated with a wide range of robotic arms and grippers, facilitating rapid deployment across diverse platforms.
    \item \emph{Improving efficiency with software-driven plug-and-play functionality}.
    Our software stack emphasizes immediate usability, requiring minimal configuration and user training.
    This design choice facilitates rapid data collection and significantly reduces operational complexity.
    Moreover, it automatically adapts to evolving hardware configurations, ensuring long-term compatibility and reliability across both current and future robotic platforms.
    \item \emph{Establishing a robust ecosystem to ensure data quality and algorithmic compatibility}.
    Our ecosystem is designed to support various imitation learning algorithms (\emph{e.g.}, Action Chunking with Transformers (ACT)~\cite{zhao2023learning} and Diffusion Policy (DP)~\cite{chi2023diffusion}) by providing essential data types, such as end-effector trajectory and joint trajectory.
    In addition, we offer tools for rapid data verification to ensure that collected datasets consistently meet the quality standards necessary for advancing manipulation capabilities.
\end{itemize}

Building upon the foundations of the original UMI, we present \textbf{FastUMI}, a redesigned system that addresses both hardware and software concerns to meet the stated objectives. 
On the \emph{hardware} side, we introduce a set of standardized, plug-and-play fingertip attachments\textemdash identical to those on the handheld device\textemdash that can be easily fitted onto a wide range of commonly used robotic grippers. 
To address the variability in robotic arm and gripper geometries, we provide a versatile, ISO-standard-compatible camera mounting solution.
By adjusting this mount, users can maintain a uniform GoPro perspective across different hardware configurations. 
This modular approach ensures consistent viewpoints, allowing models trained on handheld-collected data to be seamlessly transferred to robot-mounted scenarios.
In addition, we simplify the mechanical structures of the hand-held and robot-mounted components, enhancing overall stability and durability for extended data collection.

On the \emph{software} side, we replace the UMI's VIO-based localization with a RealSense T265 module, which integrates both visual and inertial data to provide more stable tracking in partially occluded environments.
While the GoPro can still be used for capturing high-resolution video, it is no longer relied upon as the primary visual tracking sensor.
This adjustment eliminates the necessity for separate VIO pipelines and reduces complex calibration procedures, thereby streamlining system integration and enabling rapid deployment.
In addition, we provide a robust data collection, verification, and processing pipeline designed to enhance system versatility and usability.
FastUMI supports generating two distinct types of datasets\textemdash end-effector trajectory and joint trajectory\textemdash to align with the specific input requirements of different algorithms.
Specifically, the framework accommodates two prominent categories of imitation learning algorithms\textemdash ACT and DP.
%To further address diverse manipulation scenarios, we also develop multiple custom algorithmic variants, extending the applicability and flexibility of the framework.
To tackle the unique policy-learning challenges posed by FastUMI's data\textemdash including close-up first-person perspective, variable scene geometry, and limited depth information\textemdash we develop multiple custom algorithmic variants, thereby extending the framework's applicability and flexibility across diverse manipulation scenarios.

To validate the consistency of observations and the reliability of data collection, we conduct rigorous testing, confirming that FastUMI performs similarly with the original UMI system while significantly reducing user overhead. 
Comprehensive experimental evaluations demonstrate that the redesigned system delivers an integrated, user-friendly solution that seamlessly aligns the handheld interface with robot-mounted equipment, effectively enabling efficient and scalable data acquisition for robotic learning.
We open-source over \textbf{10,000 demonstration trajectories} collected in real-world settings across 22 everyday tasks, establishing our dataset as one of the most comprehensive UMI-like collections in terms of \emph{task variety}. 
This broad range encompasses diverse manipulation scenarios and includes numerous instances of visual occlusions, which replicate real-world conditions where visual data may be intermittently obscured. 
By collecting such a diverse and challenging dataset, we not only demonstrate FastUMI's adaptability and robustness across various manipulation scenarios but also provide a valuable training resource for imitation learning algorithms.

\section{Hardware-Centric Prototype Design}\label{sec:hardware}
% The FastUMI system features a hardware-centric redesign grounded in a decoupled design philosophy to overcome key challenges. 
% This section outlines the principles that guide this redesign and provides detailed descriptions of the handheld device, the robot-mounted device, mechanisms for visual alignment, and additional optimizations that enhance system robustness and usability.
The FastUMI system embodies a hardware-centric redesign based on a decoupled design philosophy. 
This section introduces the guiding principles behind this approach and provides an overview of the hardware components.

\subsection{Hardware Design Challenges}
Aligned with the stated objectives, our hardware design must overcome several critical challenges. 
The first involves \emph{decoupling the system from specific robotic hardware}. 
By designing mechanical components that seamlessly integrate with diverse robotic arms and grippers\textemdash each varying in size, shape, and mechanical interface\textemdash we aim to minimize redesign efforts and configuration overhead. 
The second major challenge is \emph{maintaining visual consistency between handheld and robot-mounted devices}, which is essential for effective policy transfer in robotic learning algorithms. 
Given the wide range of possible gripper dimensions, preserving uniform camera perspectives across different hardware setups becomes essential. 

In addition, our design must \emph{accommodate a wider variety of robot-mounted grippers}, moving beyond the parallel-jaw restriction and thereby broadening its applicability across various robotic platforms.
Furthermore, \emph{fast deployment} is a key requirement, so we strive for a plug-and-play solution that streamlines user setup, minimizing calibration and configuration efforts to promote widespread adoption. 
Finally, \emph{ensuring high data quality} underlies the entire effort. 
Reliable hardware is crucial for obtaining accurate and consistent data, thus mitigating barriers in downstream learning tasks and enhancing overall system performance.

% \subsection{Decoupled Design Philosophy} 
% \vspace{0.5em}
% \noindent\textbf{Decoupled Design Philosophy}:
To address these challenges, FastUMI adopts a \textbf{decoupled design philosophy} that underpins its hardware architecture.
% Its core objective is to minimize dependencies on specific robotic hardware while preserving consistent, high-quality data collection. 
% By adopting a modular and standardized approach, the system is decoupled along three primary dimensions:
The system is systematically decoupled along three primary dimensions. 
Details are presented at Sections~\ref{sec:handheld} and ~\ref{sec:robot-mounted}.
\begin{itemize}
\item \textbf{Physical Decoupling}: Standardized interfaces and modular components enable seamless integration across a range of robotic platforms, eliminating the need for extensive hardware-specific modifications.
\item \textbf{Visual Consistency}: Uniform camera perspectives between handheld and robot-mounted configurations ensure that data acquired in one setting can be readily transferred to another without requiring extensive recalibration. 
This consistency allows data from human demonstrations to be directly applicable to robotic execution. 
% In practice, the handheld device employs a fixed camera configuration (thus requiring minimal alignment), whereas the robot-mounted device requires an adjustable setup due to variations in arm geometries and end-effector designs.
\item \textbf{Operational Independence}: The system incorporates self-contained tracking and sensing modules, reducing reliance on external computational frameworks and ensuring robust performance across diverse deployment scenarios.
\end{itemize}

% These principles materialize through a suite of customized hardware elements, including plug-and-play fingertip attachments and adaptable mounting mechanisms. 
% Taken together, they form a unified framework for rapid deployment, broad compatibility, and consistently high data quality.

\subsection{Handheld Device Design}\label{sec:handheld}
The handheld device (see the left subfigure in Fig.~\ref{fig:overview}) enables manual data collection for training action policies. 
It comprises three primary components:
\begin{itemize} 
\item \textbf{Fisheye Camera Module}\ding{173}:
A GoPro camera with a fisheye extension captures wide-angle images with a 155-degree field of view (FOV), significantly reducing occlusions and providing a broad perspective for robotic tasks.
This FOV is substantially larger than that of commonly used cameras such as the RealSense D435i, whose narrower field of view has proven suboptimal for first-person view (FPV) data collection in our tests.
In contrast, the wider coverage of a fisheye camera effectively captures more environmental context and enhances visual feature extraction, thereby improving policy learning outcomes.
\begin{figure}[h]
    \centering
    \includegraphics[width=0.9\linewidth]{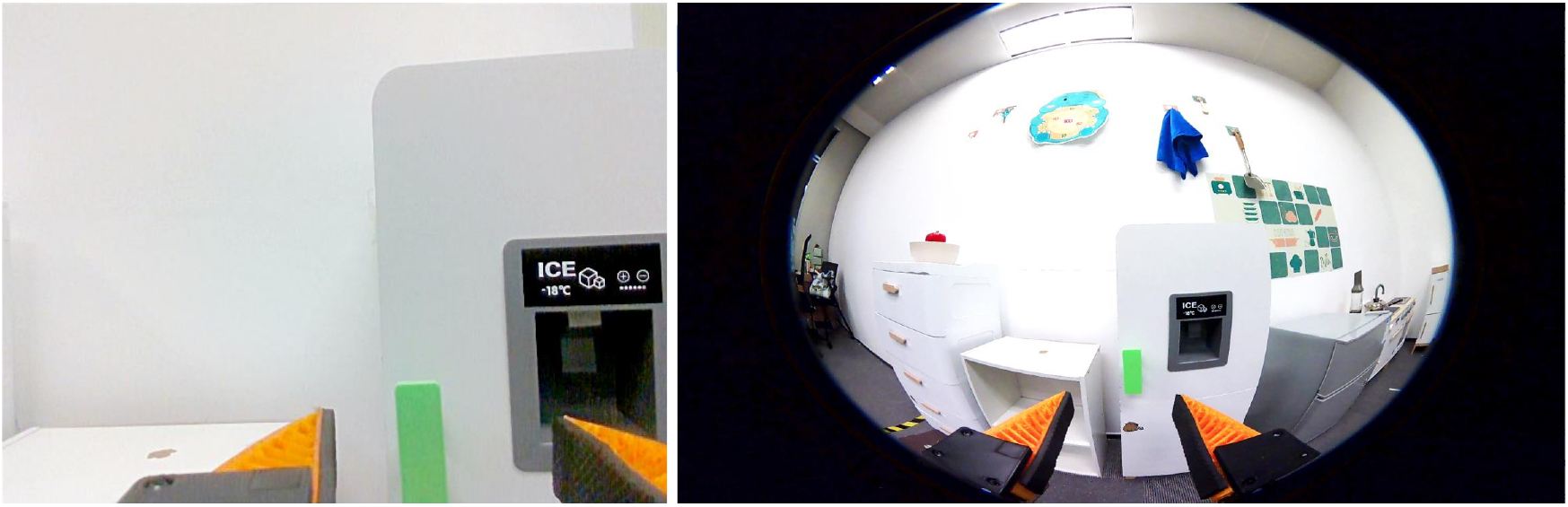}
    \caption{\textbf{Left:} D435i camera with a narrower field of view. \textbf{Right:} GoPro with a 155-degree wide-angle view.}
    \label{fig:camera}
\end{figure}
\item \textbf{Pose Tracking Module}\ding{172}:
The handheld interface lacks intrinsic joint feedback or an external motion-capture system, so we incorporate a RealSense T265 for robust tracking. 
The T265, featuring a high-performance integrated IMU, replaces UMI's visual odometry solution, eliminating complex calibration steps and enhancing usability.
% While UMI claims to handle temporary visual degradation, such as motion blur or a lack of texture details, \textbf{our experiments reveal significant limitations in scenarios involving sustained visual occlusion, such as opening cabinets or drawers.}
% In these cases, visual odometry tracking system frequently fails, leading to task interruptions. 
% In contrast, our tracking module maintains robust pose estimation under such challenging conditions. 
\emph{In our experiments, this module consistently delivers stable pose estimation across a wide range of scenarios, including those involving partial visual occlusions (e.g., opening cabinets and drawers).}
This improved hardware design accommodates a broader set of conditions, which is crucial for data collection. 
% Since the T265 has been discontinued, we have validated the RoboBaton MINI as a reliable alternative that offers comparable performance and requires no major modifications to the existing pipeline. 
% The MINI provides robust visual tracking, and a fully compatible data interface, ensuring high-quality data collection and continued availability.
Moreover, our system is \emph{robust to different camera models}, allowing smooth integration of alternative sensors without major modifications to the existing pipeline. 
For instance, we have verified that the RoboBaton MINI provides performance comparable to the T265 while maintaining a fully compatible data interface, ensuring high-quality data collection and continued availability. 
A comparison of the T265 and MINI is presented in Section~\ref{appendix_t265_mini}.

\item \textbf{Top Cover, Fingertip, and Marker}\ding{174}\ding{175}\ding{176}:
% Building upon the original UMI, we redesign the fingertip markers and top cover \emph{to enhance hardware decoupling}.
% Specifically, while the top cover is retained in the handheld device, it is intentionally excluded from the fisheye camera's field of view, acknowledging that the robot-mounted device may not incorporate a top cover.
% To achieve this, the GoPro is repositioned closer to the fingertips, which results in the markers being aligned beneath the fisheye lens.
% This adjustment, however, introduces increased distortion effects, rendering the original marker placements unsuitable for reliable detection.
% Consequently, we optimize the size and placement of the markers to minimize distortions while ensuring ease of attachment and durability.
In the original UMI design, the top cover often appears in the GoPro's field of view, \emph{preventing complete hardware decoupling}. 
To address this limitation, we reposition the GoPro closer to the fingertips and ensure the top cover remains outside the fisheye lens range, thereby accommodating setups where the cover may be absent (e.g., when mounted on a robot).
Although moving the camera closer to the fingertips naturally introduces greater image distortion, we tackle this challenge by optimizing both the size and placement of the markers.
These refinements minimize lens distortion effects, improve marker detection accuracy, and enhance durability and ease of attachment.
% Overall, this redesign boosts the flexibility and robustness of the system for a wide range of data collection scenarios, while maintaining a clear distinction between the camera module and cover components.
This redesign increases system flexibility and reliability.
\end{itemize}

In this configuration, the camera is factory-calibrated and aligned with the fingertips, requiring no further user adjustment and enabling a straightforward \emph{plug-and-play} experience.
We employ two camera modules in FastUMI, each serving a distinct function:
the T265 provides accurate pose tracking even under partial occlusions, while the GoPro delivers an expansive view crucial for environmental context capture, demonstration verification, and learning algorithm support.
Because the GoPro is not responsible for pose tracking, it can be mounted more flexibly to maintain consistent viewpoints across both handheld and robot-mounted devices,
whereas the T265 is placed in a more protected location to ensure stable pose tracking performance.
% This design ensures that, despite varying hardware platforms, the system offers both robust localization (via the T265 in the handheld version) and uniform visual framing (via the GoPro in both versions), ultimately supporting reliable, cross-scenario data collection.
% A custom limiter  ensures that the T265 remains perpendicular to the GoPro, streamlining installation without relying on complex SLAM algorithms. We also carefully align the camera's viewpoint with the gripper's fingertips to ensure consistency with the robot-mounted device.

\subsection{Robot-Mounted Device Design}\label{sec:robot-mounted}
\begin{figure}[h]
    \centering
    \includegraphics[width=0.8\linewidth]{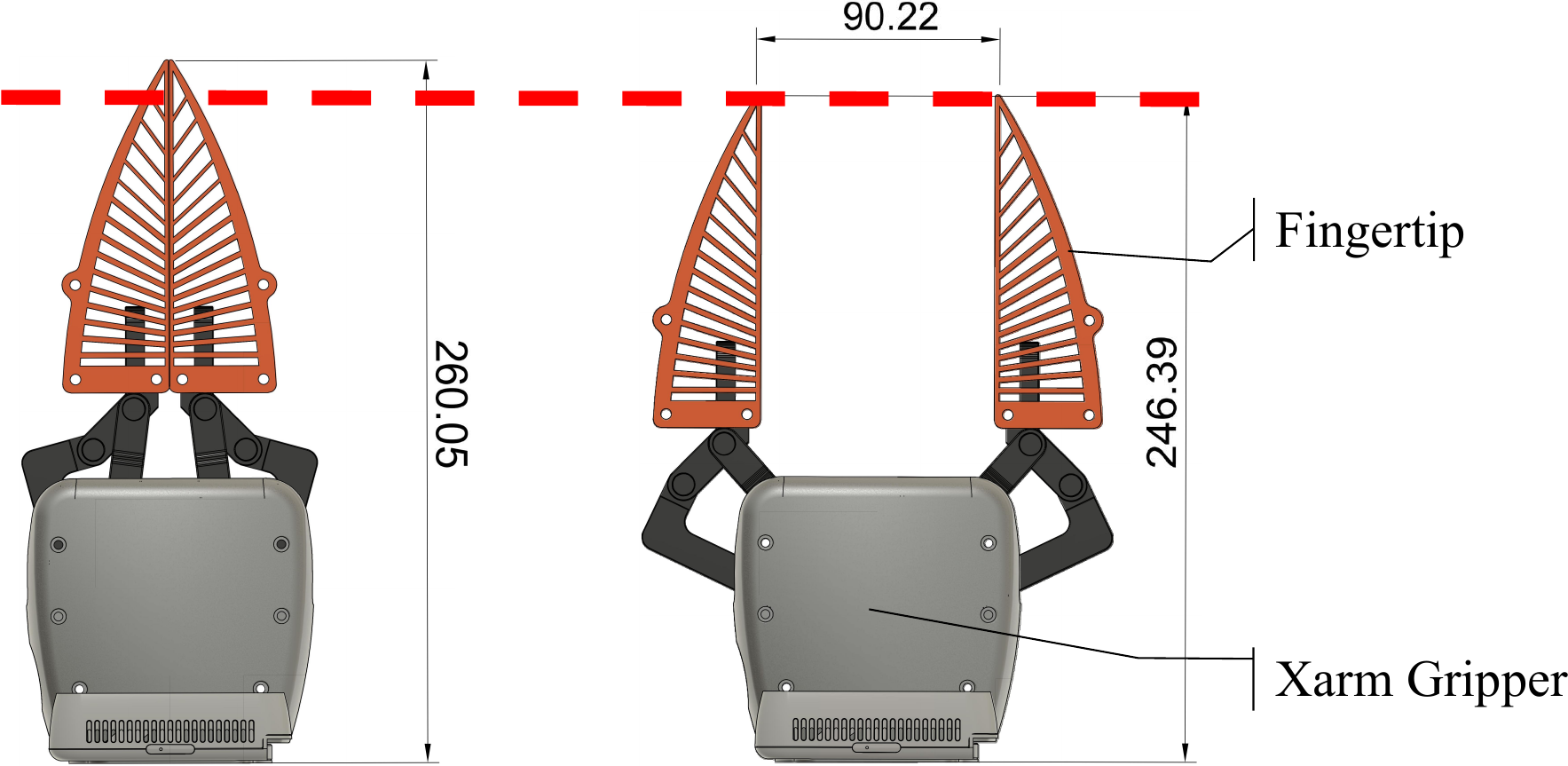}
    \vspace{-0.3em}
    \caption{Our plug-in fingertip design integrated with the xArm Gripper; The effective length of the xArm Gripper changes by approximately 1 centimeter between fully closed and open positions, potentially causing misalignment when transferring demonstrations.
    }
    \label{fig:parallel}
\end{figure}

The robot-mounted device (see the middle subfigure in Fig.~\ref{fig:overview}) follows  the same design principles as its handheld counterpart but is \textbf{engineered for broad compatibility with a wide range of robotic arms and grippers.} 
Unlike the handheld configuration, the robot-mounted device does \emph{not} include a T265 camera.
% instead, it leverages the robot's joint encoders and kinematic model for pose tracking. 
% GoPro is employed to uphold visual consistency. 
Its main components include:
\begin{itemize}
\item \textbf{Flange Plate}\ding{180}: Designed in compliance with ISO standards, the flange plate is compatible with a wide range of robotic arms, ensuring seamless integration and significantly reducing setup time.

\item \textbf{Plug-in Fingertip}\ding{181}: Outwardly identical to the attachments used in the handheld device, these modules are \emph{internally contoured to accommodate varying gripper shapes while preserving uniform external interaction points}, thereby facilitating effective policy transfer. 
Interchangeable fingertip modules establish standardized physical interaction points, supporting compatibility with various robotic and handheld grippers. 
To accommodate a wide range of robotic grippers, we design five customized fingertip attachments (\emph{e.g.}, the xArm gripper and Robotiq 2f-85) based on commonly used grippers in open-source datasets such as Open X-Embodiment~\cite{o2024open}, covering over 90\% of the grippers in these datasets.
Although not all gripper types are yet supported, our design can be readily adapted as needed.
Fig.~\ref{fig:parallel} illustrates our fingertip design integrated with the xArm Gripper.

\item \textbf{Adjustable Camera Mounting Structure}\ding{177}\ding{178}\ding{179}: Modular extension arms facilitate precise alignment of the GoPro with the gripper's fingertips, ensuring consistent viewpoints across different robot setups. 
This structure comprises two key parts:
1) \emph{GoPro Robotic Mount}\ding{177} serves as the primary attachment point for the GoPro. 
2) \emph{GoPro Extension Arm}\ding{178}\ding{179} enable both lateral positioning (indicated by the blue arrow) and vertical positioning (indicated by the red arrow) to align the camera with the robot gripper, as demonstrated in Fig.~\ref{fig:overview}.
Standardized male-female interfaces allow sequential connections of extension arms, providing adjustable length with minimal vibration (tested up to three extensions).
By adjusting the extension arm, users can replicate the handheld device's camera perspective, even when grippers vary widely in size or shape. 
Insertable fingertip extensions further ensure consistent viewpoints across heterogeneous hardware configurations.  
\end{itemize}

\begin{figure}[h]
    \centering
    \includegraphics[width=1\linewidth]{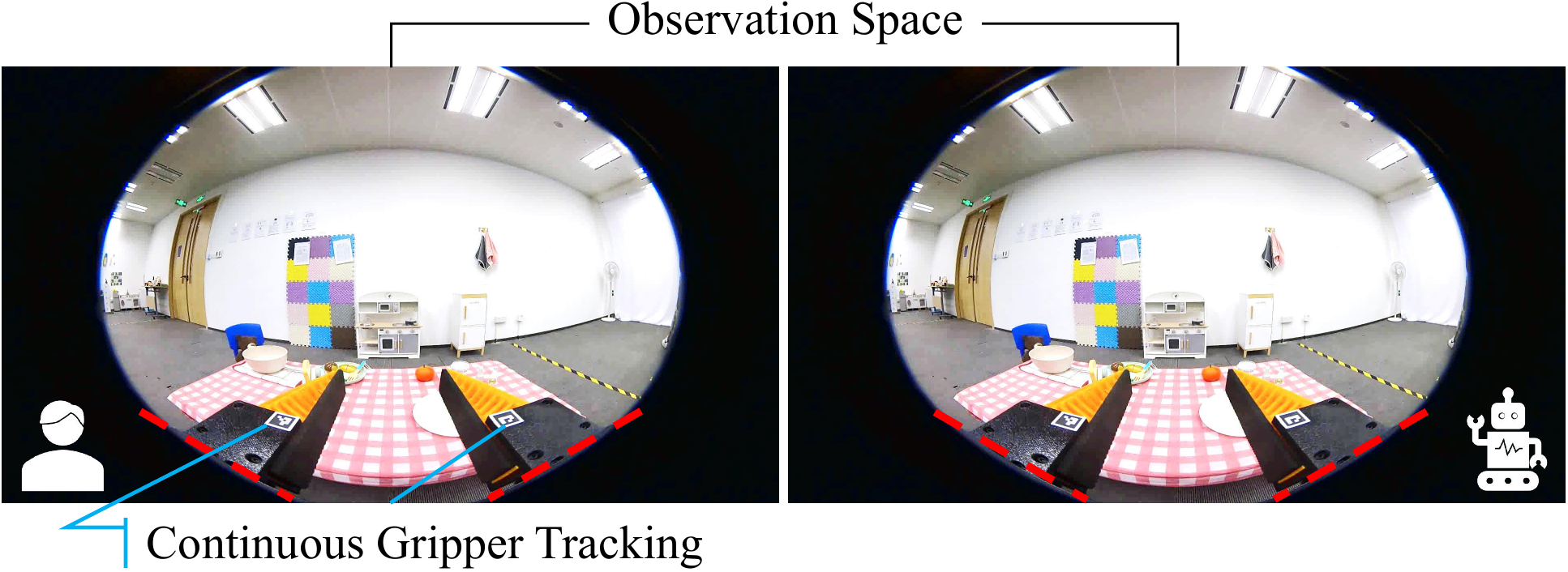}
    \vspace{-1em}
    \caption{
    Visual alignment between the handheld device (\textbf{Left}) and the robot-mounted device (\textbf{Right}). 
    The two views demonstrate the consistent positioning of the GoPro's fisheye lens image, with the bottom of the gripper's fingertips aligned to the red dashed lines.
    }
    \label{fig:alignment}
\end{figure}

\vspace{0.5em}
\noindent\textbf{Visual Alignment}:
To ensure visual consistency between the handheld and robot-mounted devices, we adopt a straightforward rule:
\emph{``The bottom of the GoPro's fisheye lens image aligns with the bottom of the gripper's fingertips.''}, as illustrated in Fig.~\ref{fig:alignment}.
This standard viewpoint ensures that all users capture nearly identical observations, enhancing interoperability across different deployments.
Although alternative standards could be defined, deviating from this alignment would reduce the utility of shared datasets for broader applications.
If gripper sizes vary, our adjustable mechanical design accommodates fine-grained arm adjustments to maintain visual alignment.
In practice, the handheld device employs a fixed camera configuration, whereas the robot-mounted device requires an adjustable setup due to variations in arm geometries and end-effector designs.

Although our handheld device is a \emph{parallel-motion gripper}, many robot-mounted grippers, such as the xArm Gripper or Robotiq Gripper, do not strictly maintain parallel motion. 
For example, the xArm Gripper's effective length changes by approximately 1 centimeter as it moves between fully open and closed positions (see Fig.~\ref{fig:parallel}). 
This discrepancy in gripper motion can create mismatches in the observed camera view, especially when transferring demonstrations collected on the handheld device to different robot-mounted setups. 
To resolve this challenge, we develop a \emph{dynamic error-compensation algorithm} that compensates for gripper-specific motion differences during inference, thereby preserving consistent visual alignment between human demonstrations and robotic executions (see Section~\ref{sec:error_compensation} for details).

\subsection{Other Design Optimizations}
% Our redesign departs from the original UMI by removing mirrors on the sides of the handheld device, as experimental evaluations indicated limited benefit. 
% Eliminating these mirrors provides space for additional components (\emph{e.g.}, tactile sensors), facilitating future expansions. 
To improve the stability and durability of FastUMI, we introduce three structural enhancements:
\begin{itemize}
\item \emph{Reinforced Key Mount Structure:} Increased structural integrity to reduce vibration.
\item \emph{Carbon Fiber Components:} Strengthened material properties while minimizing weight.
\item \emph{Standardized Male-Female Interface Design}: Allowed sequential connection of extension arms to adjust length without significant vibration.
\end{itemize}

Overall, the extensive hardware-related designs ensure reliable performance during data collection and simplify hardware adjustments for users. 
Additionally, our system configuration allows for a single standardized handheld device to be shared among multiple users, while the robot-mounted device can be adapted to various grippers or robotic arms. 
This decoupled arrangement preserves uniform data collection workflows and advances our goal of making FastUMI accessible to a broader user community.

\section{Software-Focused Framework}

\subsection{Raw Data Acquisition}
%We present the raw data collection pipeline in FastUMI, detailing how we tackle the new challenges introduced by our dual-sensor hardware design (especially the inclusion of the T265).

\vspace{0.5em}
\noindent\textbf{Overall Data Acquisition Pipeline}:
FastUMI employs three main ROS nodes to record multimodal demonstration data.
First, a \emph{camera node} continuously streams wide-angle images (e.g., 1920$\times$1080 at 60 fps) from a GoPro Camera.
Second, a \emph{tracking node} provides pose estimates from the T265 sensor at a higher rate (e.g., 200 Hz).
Each pose is represented as $(x, y, z, q_x, q_y, q_z, q_w)$, where $(x, y, z)$ represents the translation vector and $(q_x, q_y, q_z, q_w)$ represents orientation in quaternion form. 
Finally, a \emph{storage node} aggregates and synchronizes these streams in an \emph{Hierarchical Data Format version 5 (HDF5)} file for subsequent processing.
Because each sensor runs independently, the system is readily extensible to accommodate additional modalities (e.g., tactile sensors) by adding corresponding ROS nodes.

FastUMI's reliance on the T265 not only improves reliability under partial occlusions but also removes the need for extensive calibrations and VIO parameter tuning, significantly accelerating deployment. 
Nonetheless, \emph{it introduces more demanding dual-sensor synchronization and drift management requirements}, as detailed below.

\vspace{0.5em}
\noindent\textbf{Data Sub-Sampling and Synchronization}:
In multisensor data fusion, differing sampling rates and data patterns often hinder precise alignment.
To address these challenges, we employ a \emph{unified ROS clock} for consistent timestamping, a \emph{multi-threaded buffering mechanism} to handle each sensor stream independently, and \emph{synchronized sub-sampling} at the greatest common frequency.
This integrated strategy ensures robust multi-modal alignment without compromising data integrity.
Such measures are necessary because certain sensors (e.g., the T265 at 200 Hz and the GoPro at 60 Hz) exhibit mismatched rates, increasing the risk of misalignment and overload. 
In practice, each sensor's data is tagged with the unified clock and routed into a dedicated thread-safe queue to prevent data loss under high throughput. 
Before each recording session, these queues are reset to maintain orderly buffers.
We then sub-sample both streams at 20 Hz\textemdash identified as the greatest common frequency between 200 Hz and 60 Hz\textemdash by retaining one in every three camera frames and pairing each retained frame with the temporally nearest T265 pose. 
This approach achieves \emph{sub-millisecond offsets}, well within half the T265's 1/200 s interval, minimizing interpolation errors and ensuring consistently synchronized data for downstream learning tasks.

\begin{figure}[h]
    \centering
    \includegraphics[width=.9\linewidth]{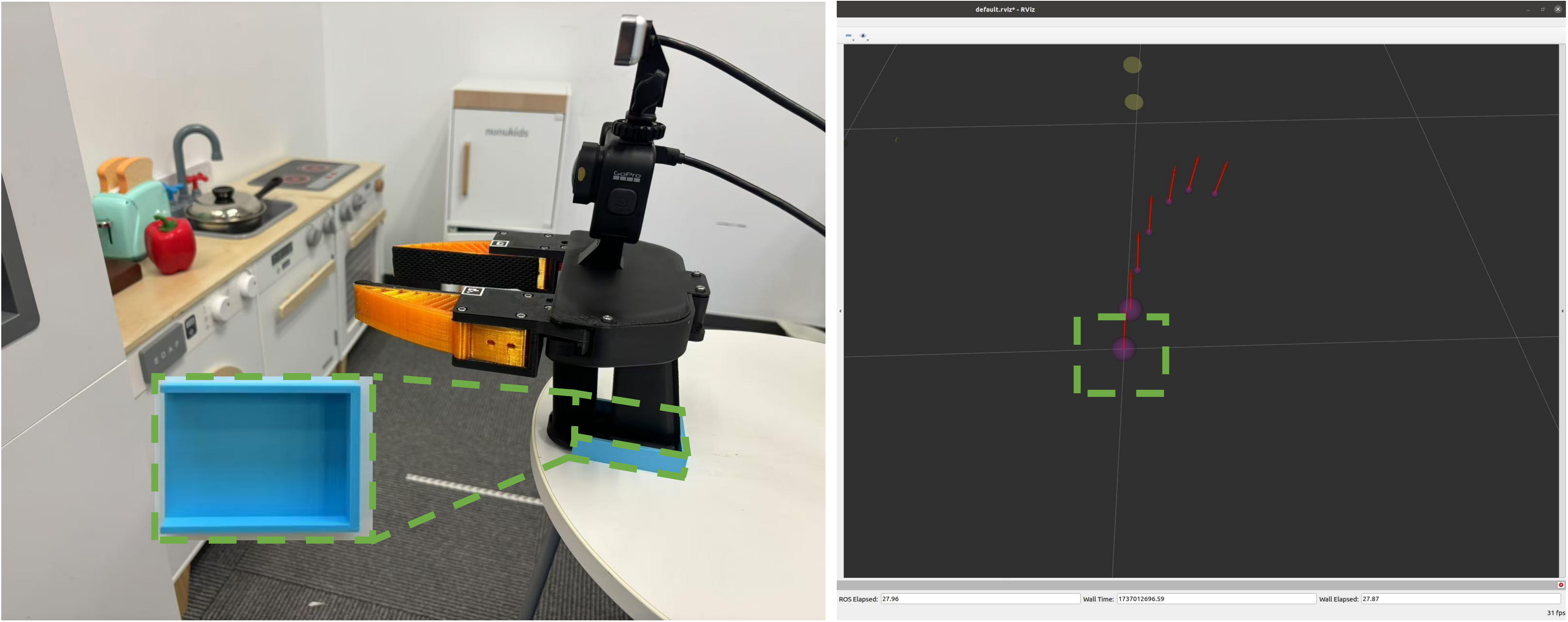}
    \vspace{-0.5em}
    \caption{\textbf{Left}: The blue 3D-printed groove on the table, serving as a clear visual reference to aid loop closure. 
    \textbf{Right}: The T265's trajectory in RVIZ, illustrating alignment with the initial reference,  highlighted as a green dashed box, after revisiting the blue groove.}
    \label{fig:drift}
\end{figure}

\vspace{0.5em}
\noindent\textbf{Accumulated Drift Correction for T265}:
Although the T265 provides robust pose estimates, it can accumulate drift during substantial motion (e.g., sudden accelerations).
% over prolonged use or substantial motion. in low-light conditions or 
% This drift arises from incremental errors in visual-inertial estimation, compounding over time into noticeable deviations.
To address this, we employ two main strategies:
\noindent\emph{1) Reinitialization.}
The simplest remedy is to restart the T265 in a stationary, predefined reference pose. 
This action resets the device's internal state and restores accurate pose tracking.
\noindent\emph{2) Loop Closure.} 
Another strategy leverages a visually distinct reference region\textemdash a blue 3D-printed groove on the table (Fig.~\ref{fig:drift} Left)\textemdash to facilitate loop closure. 
When the T265 revisits this area, it re-encounters previously mapped visual features, typically realigning the estimated trajectory with the initial reference (highlighted as a green dashed box in Fig.~\ref{fig:drift} Right) in RVIZ under minor drift. 
However, if significant misalignment persists even after returning to the marked area, loop closure is deemed ineffective, and the T265 must be reinitialized to restore accurate pose tracking.

\subsection{Raw Data Quality Assessment}~\label{sec:assessment}

Ensuring reliable demonstrations is crucial for downstream learning tasks; however, to our knowledge, no existing work fully quantifies what constitutes ``ideal'' data quality.
In practice, we enforce consistency through \emph{sensor confidence} and \emph{trajectory smoothness} checks. 
The T265 provides four discrete confidence levels\textemdash Failed, Low, Medium, and High; to avoid prolonged low-confidence data, we first validate the environment by confirming that at least 95\% of sample poses achieve \emph{High} confidence. 
Our tests indicate that \textbf{lighting conditions} notably affect the T265's performance, with dim or low-light environments often leading to reduced confidence levels and increased drift.
During actual recordings, any low-confidence pose is excluded and interpolated from neighboring frames to maintain continuity.
Meanwhile, user-defined thresholds on \emph{velocity}, \emph{acceleration}, and \emph{relative orientation} identify abrupt transitions, further refining data fidelity. 
Although these strategies cannot guarantee an absolute benchmark for data quality, they help establish rigorous collection standards and minimize errors that could propagate into subsequent modeling and inference.

\subsection{Data Preparation for Training}

\vspace{0.5em}
\noindent\textbf{Data Type Overview}:
% To accommodate different imitation learning algorithms, we provide two primary categories of trajectory data: \textbf{TCP trajectories} (both absolute and relative), and \textbf{joint trajectories}. 
To address the requirements of diverse imitation learning algorithms, we categorize trajectory data into two main types\textemdash \textbf{TCP trajectories} (both absolute and relative), and \textbf{joint trajectories}. 
% % In practice, TCP trajectories are often employed by DP, whereas ACT requires joint trajectories. 
At the software level, FastUMI facilitates seamless integration of various data formats and evolving algorithmic needs with minimal configuration. 
% This design supports a wide range of imitation learning methods while maintaining long-term scalability and extensibility. 
Below, we outline the procedures for generating these three data formats.

\begin{figure}[t]
    \centering
    \includegraphics[width=0.75\linewidth]{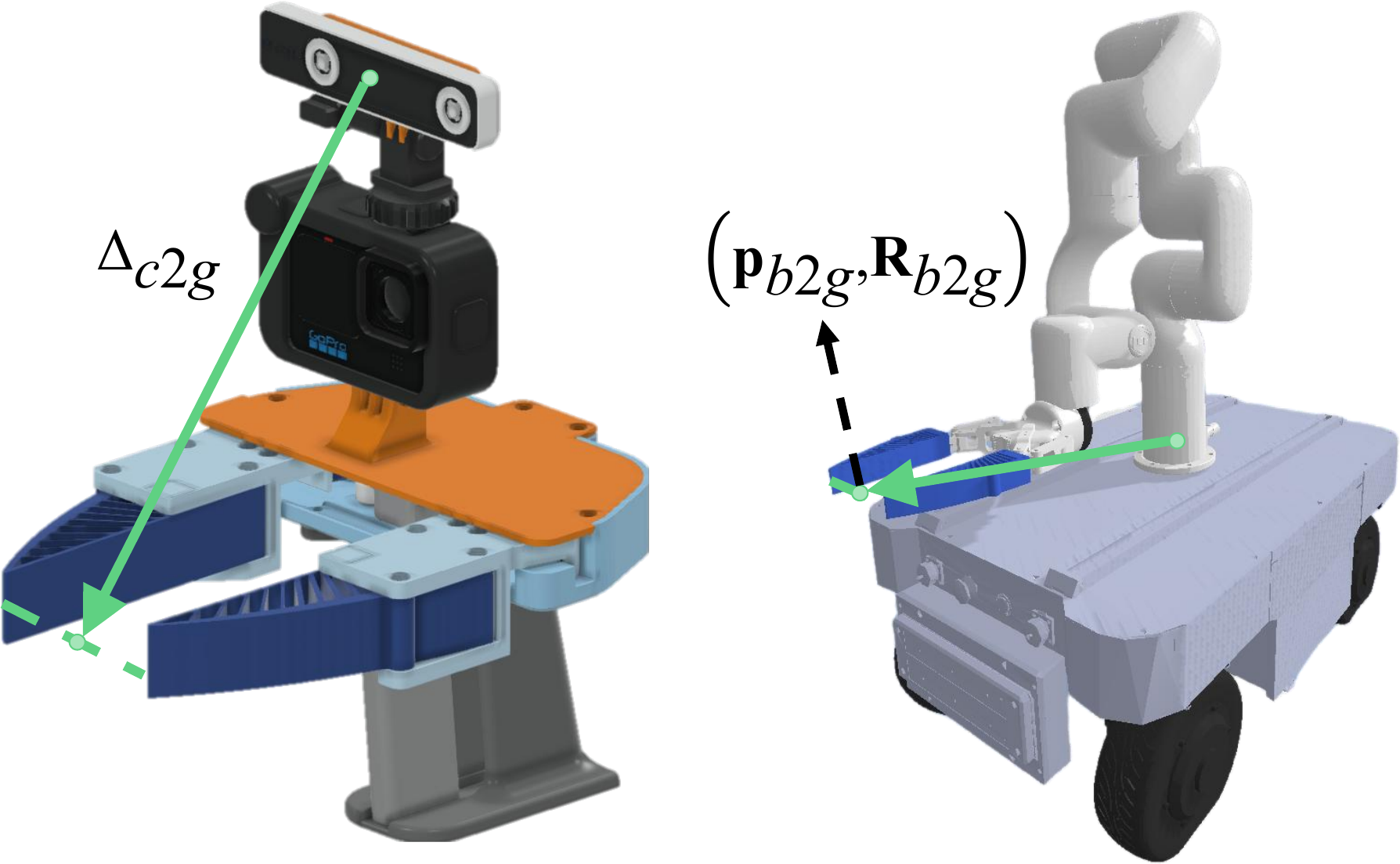}
    % \vspace{-0.5em}
    \caption{Illustration of the offset \( \Delta_{c2g} \) from the T265 center to the gripper center, and the gripper center pose \(\bigl(\mathbf{p}_{b2g}, \mathbf{R}_{b2g}\bigr)\) in the robot base frame.}
    \label{fig:transformation}
\end{figure}

\vspace{0.5em}
\noindent\textbf{Input Data and Assumption}:
The principal raw inputs are pose estimates from the T265 camera, given by \( \mathbf{p}_i \) (position) and \( \mathbf{R}_i \) (orientation) in the camera's local frame. 
The following additional information is also required: 
\begin{itemize}
    \item The \emph{known} robotic arm's Unified Robot Description Format (URDF).
    \item The \emph{known} offset \( \Delta_{c2g} \) from the T265 \underline{c}amera center to the \underline{g}ripper center (expressed in the camera frame), as shown in Fig.~\ref{fig:transformation} (Left).
    \item The \emph{known} pose \(\bigl(\mathbf{p}_{b2g}, \mathbf{R}_{b2g}\bigr)\) of the gripper center in the robot \underline{b}ase frame, as shown in Fig.~\ref{fig:transformation} (Right), where \(\mathbf{p}_{b2g}\) is the position, and \(\mathbf{R}_{b2g}\) is the rotation.
\end{itemize}
We assume that the hand-held device motion precisely mirrors that of the robot end-effector.
Under these conditions, the following trajectories can be derived: 1) \emph{Absolute TCP trajectories}, 2) \emph{Relative TCP trajectories}, and 3) \emph{Absolute joint trajectories}.
Details are outlined below.

\vspace{0.5em}
\noindent\textbf{Absolute TCP Trajectory}:
To compute this trajectory, the T265 coordinate system is first aligned with the robot base frame. 
At each timestamp \(i\), the T265 provides \(\bigl(\mathbf{p}_i, \mathbf{R}_i\bigr)\), describing the camera's motion relative to its initial pose. 
The camera's absolute pose in the robot base frame is given by:
\begin{align}
\mathbf{p}_\text{cam}^{(i)} &= \mathbf{p}_\text{b2g}
  \;+ \mathbf{p}_i \;-\; \mathbf{R}_\text{b2g} \,\Delta_{c2g}, \\[4pt]
\mathbf{R}_\text{cam}^{(i)} &= \mathbf{R}_\text{base}
  \;\cdot\; \mathbf{R}_i.
\end{align}
The absolute TCP pose \(\bigl(\mathbf{p}_\text{ee}^{(i)}, \mathbf{R}_\text{ee}^{(i)}\bigr)\) is then obtained by incorporating the camera-to-gripper offset \( \Delta_{c2g} \):
\begin{align}
\mathbf{p}_\text{ee}^{(i)} &= \mathbf{p}_\text{cam}^{(i)} 
  \;+\; \mathbf{R}_\text{cam}^{(i)} \,\Delta_{c2g}, \\[3pt]
\mathbf{R}_\text{ee}^{(i)} &= \mathbf{R}_\text{cam}^{(i)}.
\end{align}
The resulting sequence \(\bigl\{(\mathbf{p}_\text{ee}^{(i)}, \mathbf{R}_\text{ee}^{(i)})\bigr\}\) yields the absolute TCP trajectory in the robot base frame.

\vspace{0.5em}
\noindent\textbf{Relative TCP Trajectory}:
This trajectory is formed from consecutive absolute TCP frames. 
For adjacent frames \( i \) and \( i+1 \), with absolute poses \(\bigl(\mathbf{p}_\text{ee}^{(i)}, \mathbf{R}_\text{ee}^{(i)}\bigr)\) and \(\bigl(\mathbf{p}_\text{ee}^{(i+1)}, \mathbf{R}_\text{ee}^{(i+1)}\bigr)\), the relative transforms are:
\begin{align}
\mathbf{p}_\text{rel}^{(i)} &= \mathbf{p}_\text{ee}^{(i+1)} 
  \;-\; \mathbf{p}_\text{ee}^{(i)}, \\[3pt]
\mathbf{R}_\text{rel}^{(i)} &= \bigl(\mathbf{R}_\text{ee}^{(i)}\bigr)^{-1}
  \;\cdot\; \mathbf{R}_\text{ee}^{(i+1)}.
\end{align}
This formulation removes dependence on a global reference, facilitating more uniform data distributions and improving generalization when the base pose varies.

\vspace{0.5em}
\noindent\textbf{Absolute Joint Trajectory:}
% Some imitation learning methods require joint trajectories. 
To obtain it, inverse kinematics (IK) is solved for each absolute TCP pose \(\bigl(\mathbf{p}_\text{ee}^{(i)}, \mathbf{R}_\text{ee}^{(i)}\bigr)\) using the robot's URDF, typically via an iterative solver.
To maintain continuity, the solution at frame \( i \) serves as the initial guess for frame \( i+1 \). 
If the URDF only extends to the flange, the flange-to-gripper offset is accounted for in the IK computations to ensure accurate joint solutions. 
% The resulting sequence is the \emph{absolute joint trajectory} required by algorithms such as ACT.

\vspace{0.5em}
\noindent\textbf{Continuous Gripper Width Computation:}
We propose a marker-based method that \emph{decouples software from the underlying mechanical structure}, thereby facilitating compatibility with diverse gripper designs. 
% Unlike the original UMI approach\textemdash which applied fisheye lens correction and included various parallel-gripper parameters\textemdash 
Specifically, we measure the pixel distance between ArUco markers~\cite{GARRIDOJURADO20142280} on the gripper jaws and map it linearly to the gripper's physical opening width.
This approach obviates rigid hardware dependencies, reducing design constraints and streamlining integration of new or differently sized grippers.

In our setup, we attach two ArUco markers to the gripper and define two hyperparameters: \(d_{\max}\) and \(d_{\min}\). 
These values represent the pixel distances between the markers at the gripper's maximum and minimum openings, respectively. 
For each image frame, we detect the markers and compute the pixel distance \(d\). 
If only one marker is identified, we estimate \(d\) by mirroring the known marker about the gripper's central axis; if no markers are detected, an imputed value is inserted to maintain continuity. 
Consequently, each frame is guaranteed a valid marker distance.
Finally, the physical gripper width, denoted as $W$, is determined by normalizing the measured distance with respect to \(d_{\max}\) and \(d_{\min}\), then scaling by \(G_{\max}\), which denotes the jaws' maximum physical opening:
\begin{equation}
W = \frac{d - d_{\min}}{d_{\max} - d_{\min}} \times G_{\max}.
\vspace{1em}
\end{equation}

% \[
% \text{$W$} \;=\;
%   \frac{d - d_{\min}}{\,d_{\max} - d_{\min}\,}
%   \;\times\;
%   G_{\max}.
% \]

% By using pixel-based distances instead of a fixed mechanical model, this method greatly increases hardware flexibility and lowers complexity when adapting to alternative gripper configurations.

\section{Algorithmic Adaptations for FastUMI}\label{sec:algorithm}

\begin{figure*}[t]
    \centering
    \includegraphics[width=.9\linewidth]{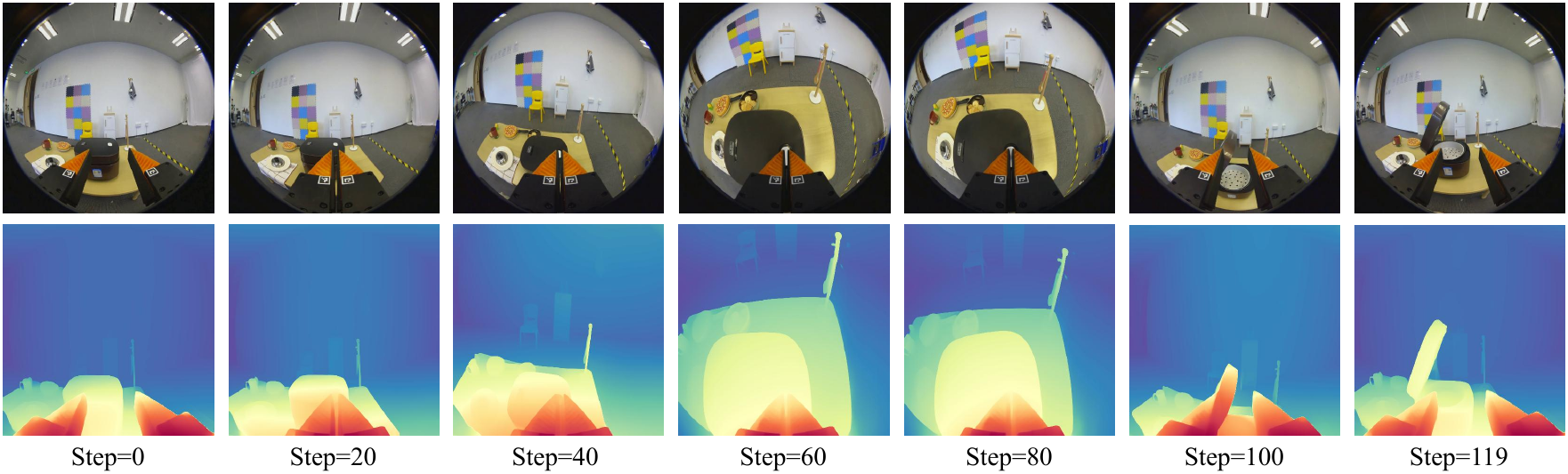}
    \vspace{-0.5em}
    \caption{Illustration of the depth-mapping process from Step 0 to Step 119. 
    The top row displays the cropped GoPro frames (rectified from their original circular format), and the bottom row presents corresponding depth estimations produced by Depth Anything V2.}
    \label{fig:dp_depth}
\end{figure*}

\subsection{Motivation: Beyond Hardware Decoupling}
In earlier sections, we introduce how FastUMI attains hardware decoupling.
% by removing dependencies on specific robot components
This design choice lowers the cost and complexity of system deployment across heterogeneous platforms, including handheld and robot-mounted configurations.
However, while hardware decoupling supports seamless data acquisition across various setups, it does not by itself address the \textbf{distinct policy-learning challenges arising from FastUMI's data distributions} (details in Section~\ref{sec:umi_data}).
% FastUMI's data, characterized by close-up first-person perspective, variable geometry and scene layouts, and limited depth information, introduces further complexities in learning robust policies.  
Hence, hardware decoupling alone cannot fully realize FastUMI's potential. 
% Algorithmic refinements are crucial to capitalize on the broader range of data captured under diverse configurations.
% By integrating specialized learning approaches with FastUMI, our goal is to bridge this gap, ensuring efficient multi-platform deployment alongside consistently high policy performance.
By incorporating data-driven refinements into baseline algorithms to accommodate FastUMI's unique data characteristics, we enable efficient multi-platform deployment with consistently high performance while also \emph{laying the groundwork for more advanced methods in the future}.

\subsection{Data Challenges with FastUMI}~\label{sec:umi_data}
Compared to conventional third-person or fixed-base perspectives, FastUMI's wrist- or handheld-mounted viewpoints introduce several distinct data characteristics:
\begin{itemize}
    \item \textbf{Close-up First-Person Perspective}:
    Cameras positioned near the end-effector capture detailed manipulation cues but offer \emph{limited visibility of the full robotic arm}, increasing dependence on priors to maintain kinematic feasibility.
    \item \textbf{Variable Geometry and Scene Layout}:
    FastUMI's hardware-agnostic design generates heterogeneous data across different arm configurations, base frames, and environments, complicating efforts to achieve consistent policy learning.
    \item \textbf{Limited Depth Information}:
    Single-view fisheye images lack explicit three-dimensional spatial cues, making precise depth estimation difficult. 
    Tasks demanding accurate positioning, including object alignment and gripper closure, are particularly vulnerable to errors when depth signals are absent.
\end{itemize}

To address these challenges, we present adaptations for two primary imitation learning algorithms: ACT and DP. 
These enhancements promote robust policy execution, ensure kinematic feasibility, and integrate depth-awareness for tasks that require higher precision.

\subsection{Enhanced ACT for First-Person Perspectives}
The standard ACT predicts \emph{absolute joint trajectories} for the robotic arm, performing effectively in third-person or fixed-camera scenarios.
However, under FastUMI's first-person wrist-mounted perspective, large portions of the robotic arm remain unseen, making ACT susceptible to producing \textbf{illicit joint configurations} during inference. 
These configurations can exhibit extreme end-effector orientations that violate kinematic constraints or diverge substantially from demonstration trajectories.
Consequently, we introduce two targeted refinements to the original ACT to address visibility limitations inherent in first-person data.

\vspace{0.5em}
\noindent\textbf{1) Smooth-ACT: Local Temporal Smoothing}.
To address abrupt or infeasible joint transitions, we introduce Smooth-ACT, which enhances action continuity by incorporating a Gated Recurrent Unit (GRU) layer on top of the Transformer decoder~\cite{dey2017gate}.
While the Transformer captures global spatiotemporal patterns, the GRU refines local continuity, smoothing sudden deviations between successive frames.
During training, two action sequences are produced: $\hat{a}$ from the Transformer decoder and $\hat{a}_{\mathrm{GRU}}$ from the GRU layer. 
Both are compared against ground truth actions with the loss function $\mathcal{L}$:
\begin{equation}
\mathcal{L} 
= \|\hat{a} - a\|_1 
+ \|\hat{a}_{\mathrm{GRU}} - a\|_1 
+ \lambda \,\mathrm{KL}\bigl(\mu, \log \sigma^2\bigr),\
\end{equation}
where $\mathrm{KL}\bigl(\mu, \log \sigma^2\bigr)$ regularizes model outputs for stability. 
This hierarchical setup preserves the Transformer's capacity for global attention while enforcing local smoothing, thereby reducing kinematically invalid actions.

\vspace{0.5em}
\noindent\textbf{2) PoseACT: End-Effector Pose Prediction}.
% We replace absolute joint predictions with an end-effector (TCP) pose representation\textemdash encompassing both absolute and relative trajectories\textemdash to form a variant called PoseACT.
Beyond mitigating trajectory discontinuities, we further enhance ACT's robustness by introducing PoseACT, a variant that replaces absolute joint predictions with an end-effector (TCP) pose representation.
This formulation incorporates both absolute and relative motion trajectories, offering two key benefits:
\begin{itemize}
    \item \textbf{Platform Independence}: 
    Expressing actions in terms of local end-effector movement reduces sensitivity to base-frame or arm-geometry variations, facilitating multi-platform policy transfer.
    \item \textbf{Numerical Stability}:
    Relative trajectories generally show less variability, mitigating outlier effects and improving generalization to novel configurations.
\end{itemize}
During inference, the policy outputs relative poses, which are subsequently mapped back to absolute joint angles via the robot's kinematic model. 
Our evaluations suggest this base-agnostic approach increases policy robustness and minimizes extreme joint commands, especially under limited first-person observations.

\subsection{Depth-Enhanced Diffusion Policy}
% To evaluate the usability of the data collected by FastUMI, we applied it to deploy a diffusion policy tailored to our robotic platform. 
% Following the same framework as UMI (relative pose and latency matching specific to xArm 6), we observed promising initial results. 
We apply DP from the original UMI (which includes relative TCP trajectory prediction and latency matching) to our robotic platform, and observe promising initial results.
However, we identify a limitation: \textbf{the DP struggle with tasks requiring high precision in depth estimation.} 
For instance, it occasionally fails to accurately reach the target or prematurely closed the grippers. 
This reveals the inadequacy of the current policy when operating without depth information, especially in scenarios where precise spatial reasoning is essential.

To address this issue, we incorporate depth information to improve the original DP, resulting in a variant, called \textbf{Depth-Enhanced DP}.
While existing works incorporating depth into DP often rely on dedicated sensors to capture real-time depth data~\cite{ze20243d,ke20243d}, we aim to explore a \emph{more lightweight and efficient approach} without adding hardware complexity or additional costs.
Specifically, we adopt a post-processing strategy to generate depth maps. 
Using the open-source depth estimation tool Depth Anything V2~\cite{yang2024depthv2}, we supplement each frame in our dataset with corresponding depth maps.
However, our images, which include significant black margins, negatively impact the performance of Depth Anything V2. 
To overcome this challenge, we pre-process the images by cropping them to retain only the rectangular regions inscribed within the circles and resizing them to 448 $\times$ 448, as shown in Fig.~\ref{fig:dp_depth}.
This ensures both high-quality depth maps and a focus on the operational areas associated with the grippers. Furthermore, the RGB images are also cropped along the black margins on both sides, followed by resizing to 448 $\times$ 448, in order to enhance the accuracy of image observation.
% Our experiments revealed that depth information primarily aids in understanding the spatial relationship between the grippers and the objects being manipulated, making it indispensable for precision-intensive tasks. 

During training, we expand the single-channel depth data into three-channel pseudo-color depth maps, which are then encoded alongside the RGB images using the same ViT-based CLIP visual encoder (ViT-Base Patch 16, input 224$\times$224)~\cite{radford2021learning} respectively. 
The resulting embeddings are concatenated and used for downstream processing.
For real-time inference during the diffusion policy rollout, we implement Depth Anything V2 with its large pre-trained model~\cite{yang2024depthv2}, achieving an inference frequency of \textbf{20 Hz} on an RTX 4090 GPU. 
% As shown in Table [], incorporating depth information into UMI-like data significantly alleviated the depth insensitivity issue encountered with current mainstream policies that rely solely on the first wrist viewpoint. 
This improvement is achieved without the need for additional sensors or multi-view camera setups, providing a practical and efficient solution to enhance policy performance in precision-critical applications.

\subsection{Dynamic Error-Compensation Algorithm}\label{sec:error_compensation}
Non-parallel-jaw grippers on robotic arms can shift the TCP as the jaws close.
Because the jaws move inward, the effective TCP often translates along the gripper's local $Z$-axis, leading to misalignment in tasks that require fine precision, such as picking up small objects.
To mitigate these shifts, we propose a \emph{dynamic error-compensation algorithm} that adjusts the commanded TCP in real time.
%guided by visual feedback on the gripper's opening width
It includes two stages.

\vspace{0.5em}
\noindent\textbf{Stage 1: Compensation Distance}.
Let $W(i)$ be the measured gripper width at frame $i$, and denote the gripper's maximum width by $W_{\max}$.
We define a compensation distance $d(i)$ to counteract TCP displacement caused by non-parallel jaws. 
Let $d_{\mathrm{close}}$ be the maximum compensation distance when the gripper is fully closed, and $d_{\mathrm{open}}$ be the minimum distance when it is fully open. 
We then compute
\begin{equation}
d(i) = d_{\mathrm{close}} - \frac{d_{\mathrm{close}} - d_{\mathrm{open}}}{W_{\max}} W(i).
\end{equation}
As $W(i)$ decreases, the end-effector is shifted further along the negative $Z$-axis of TCP frame, thereby offsetting the forward motion introduced by closing gripper jaws.

\vspace{0.5em}
\noindent\textbf{Stage 2: Pose Correction}.  
Let $\mathbf{p}_\text{ee}^{(i)}$ and $\mathbf{R}_\text{ee}^{(i)}$ be the desired TCP position and orientation at frame $i$, respectively. 
The rotation matrix $\mathbf{R}_\text{ee}^{(i)}$ defines the TCP coordinate frame's orientation relative to the robot's base frame.
To determine the direction of displacement, we first extract the TCP frame's local $Z$-axis, expressed in the base coordinate frame:
\begin{equation}
\mathbf{z}_{\text{axis}}^{(i)} \;=\; 
\mathbf{R}_\text{ee}^{(i)} \,\hat{\mathbf{e}}_z,
\end{equation}
where $\hat{\mathbf{e}}_z = [0,\,0,\,1]^\mathsf{T}$ is the local $Z$-axis of the TCP coordinate frame. 
The corrected TCP position $\mathbf{p}_\text{ee}'^{(i)}$ in the base coordinate frame is then computed as:
\begin{equation}
\mathbf{p}_\text{ee}'^{(i)} 
\;=\; \mathbf{p}_\text{ee}^{(i)} 
\;-\; d(i)\,\mathbf{z}_{\text{axis}}^{(i)}.
\end{equation}
Finally, inverse kinematics (IK) is solved using the corrected TCP position $\mathbf{p}_\text{ee}'^{(i)}$, while maintaining the original orientation $\mathbf{R}_\text{ee}^{(i)}$, yielding the joint vector $\boldsymbol{\theta}^{(i)}$:
\begin{equation}
\boldsymbol{\theta}^{(i)} \;=\; 
\mathrm{IK}\bigl(\mathbf{p}_\text{ee}'^{(i)},\,\mathbf{R}_\text{ee}^{(i)}\bigr).
\end{equation}

\begin{figure*}[t]
    \centering
    \includegraphics[width=1\linewidth]{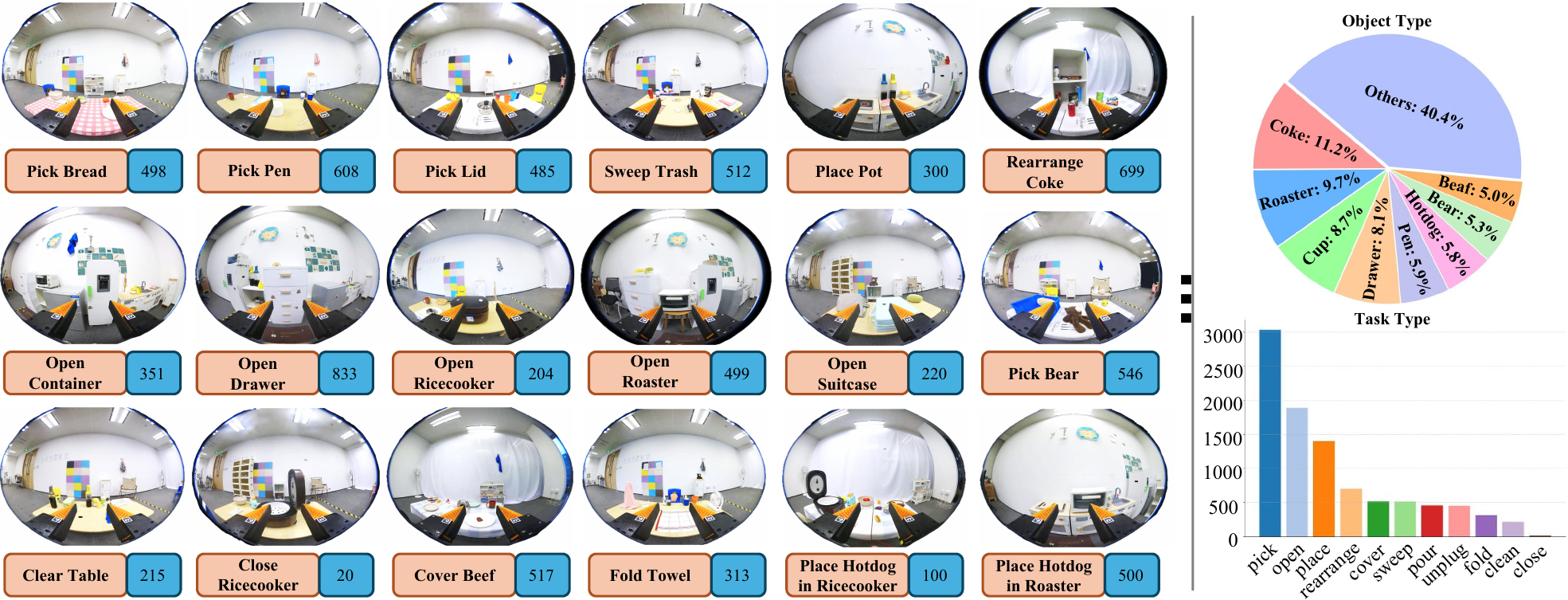}
    \caption{\textbf{Left}: Representative GoPro frames from the FastUMI dataset, with \emph{orange} labels indicating specific tasks and \emph{blue} numerals showing demonstration counts. 
    \textbf{Right}: Two distribution plots\textemdash the \textbf{top} plot depicts the proportion of various tasks in the dataset, while the \textbf{bottom} plot shows the distribution of manipulation skills.
    }
    \label{fig:dataset}
\end{figure*}

\section{Open-Source Dataset}

\subsection{Dataset Overview}
We present the FastUMI Dataset, consisting of \textbf{10,000} demonstration sequences, each containing synchronized GoPro video and end-effector trajectories captured in domestic settings.
The dataset covers \textbf{22} tasks, \textbf{19} object categories, and \textbf{12} distinct manipulation skills, with each demonstration lasting approximately \textbf{6-12} seconds, where most demonstrations are 9 seconds. 
Fig.~\ref{fig:dataset} (Left) illustrates representative frames from selected collection environments; orange text boxes denote specific tasks, while blue numerals indicate corresponding demonstration counts. 
Fig.~\ref{fig:dataset} (Right) provides two distribution plots: the upper plot details task-level proportions, whereas the lower plot illustrates the breakdown of manipulation skills (e.g., pick, open, etc.). 

\subsection{Dataset Acquisition Process}
The dataset is collected by \textbf{five} operators using \textbf{three} FastUMI devices, ensuring diversity in user interactions and environmental contexts.
Each recorded task involves a fixed target object (e.g., a specific drawer or container), while the surrounding background (e.g., table clutter, lighting) is randomized to introduce variability.
During acquisition, operators utilize RVIZ for real-time visualization, enabling verification of the T265 sensor output and ensuring high-quality demonstrations.
We enforce a quality-assurance protocol by continuously monitoring critical metrics (e.g., T265 tracking confidence) and discarding or re-recording sequences affected by sensor drift.

\begin{figure*}[t]
    \centering
    \includegraphics[width=.95\linewidth]{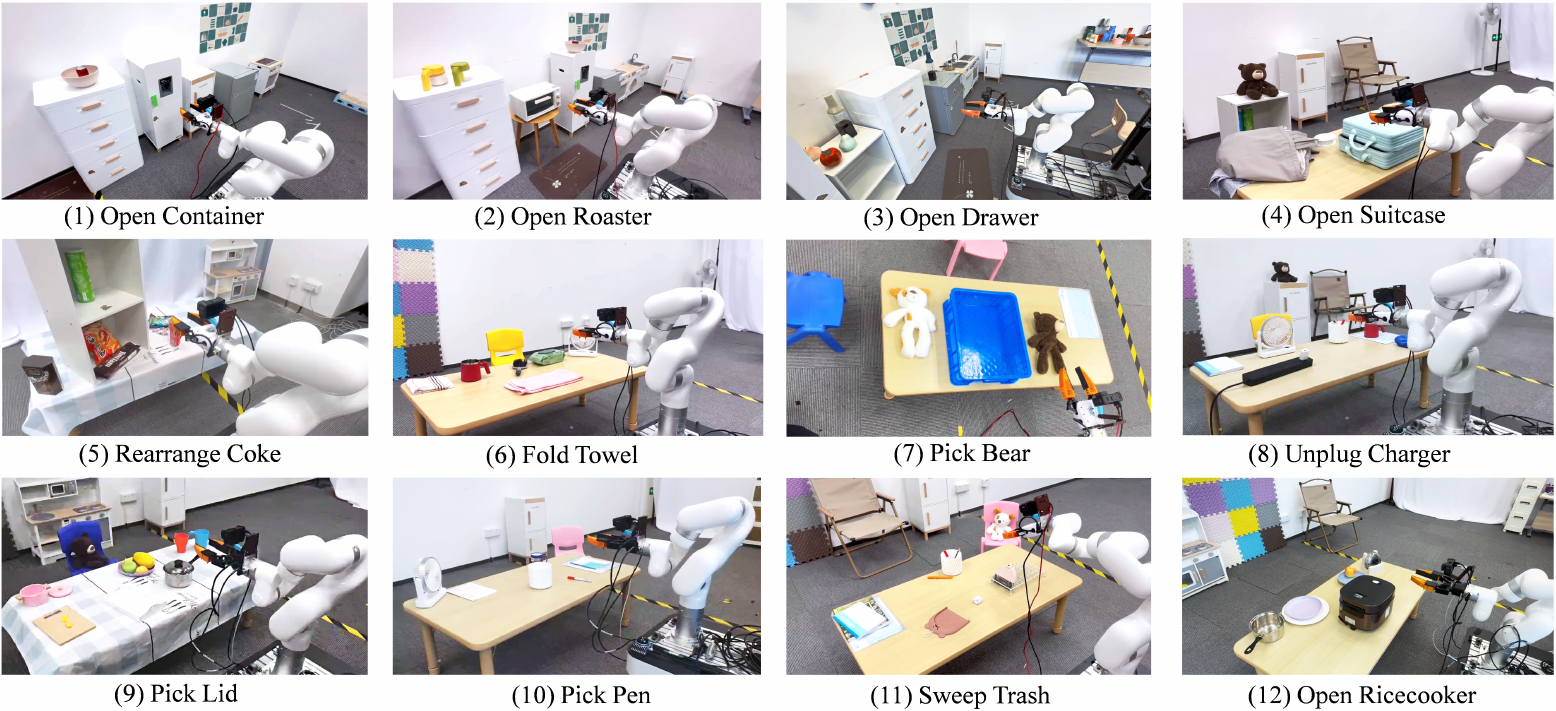}
    \caption{Twelve tasks used for policy inference evaluation, covering a broad range of real-world manipulation challenges. These include \textbf{hinged operations} (Tasks 1–4), \textbf{pick-and-place activities} (Tasks 5–10), \textbf{pick-push manipulation} (Task 11), and \textbf{button press actions} (Task 12), providing a comprehensive benchmark for the proposed system.}
    \label{fig:task}
\end{figure*}

\subsection{Dataset Storage and Format}
All raw sensor data are initially recorded locally before undergoing post-processing.
To support various imitation learning and control paradigms, we provide multiple data representations\textemdash most notably, joint trajectories and TCP trajectories.
Each demonstration is stored in a dedicated HDF5 file, encapsulating both observations (e.g., images, tracked poses) and actions (e.g., gripper commands) within a unified dataset.
For broader compatibility, we also provide scripts to convert HDF5 files into Zarr format, which maintains a hierarchical structure while offering greater flexibility in storage backends, chunking, compression, and parallel access.
Detailed specifications of the dataset schema and file organization are provided in Section~\ref{appendix_dataset}.

\section{System Evaluation}
We assess our system across four primary dimensions: \textbf{1) Data Quality}, \textbf{2) Baseline Performance}, \textbf{3) Algorithmic Enhancements}, and \textbf{4) Additional Factors}, detailed in Sections~\ref{sec:dataquality}, \ref{sec:baseline}, \ref{sec:enhancement}, and \ref{sec:additional}, respectively. 
Initially, we evaluate the reliability of the T265 and MINI modules in pose tracking to ensure that the collected data meets the prerequisites for downstream applications. 
Subsequently, we demonstrate the system's effectiveness through extensive experiments encompassing \textbf{a variety of manipulation tasks}, thereby highlighting its robustness using baseline ACT and DP methodologies. 
We then quantify the improvements in task success rates attributable to our algorithmic refinements. 
Finally, we examine additional variables, such as the utilization of fisheye cameras and the size of the training dataset, to evaluate their influence on policy performance.

Prior to presenting the experimental results, we introduce the \textbf{12 tasks} employed for policy inference evaluation, as illustrated in Fig.~\ref{fig:task}. 
These tasks are designed to encompass a broad spectrum of real-world manipulation challenges, including hinged operations and pick-and-place activities, thereby providing a comprehensive benchmark for assessing the proposed system. 
Unless otherwise specified, all experiments are conducted using an xArm 6 robotic platform.

\subsection{Data Quality}\label{sec:dataquality}

\begin{table*}[h]
\centering
\scriptsize
\renewcommand{\arraystretch}{1}
\setlength{\tabcolsep}{5pt}
\caption{Error analysis of trajectories for different tasks (values in mm).}
\label{tab:trajectory_errors}
\begin{tabular}{llcccccccccc}
\toprule
\textbf{Pose Tracking Module} & \textbf{Task} & \textbf{Traj 1} & \textbf{Traj 2} & \textbf{Traj 3} & \textbf{Traj 4} & \textbf{Traj 5} & \textbf{Traj 6} & \textbf{Traj 7} & \textbf{Traj 8} & \textbf{Traj 9} & \textbf{Traj 10} \\
\midrule
\textbf{RealSense T265} & Pick Cup & 11 & 10 & 12 & 11 & 11 & 12 & 11 & 10 & 7  & 10 \\
\cmidrule{2-12}
& Open Container & 19 & 16 & 18 & 17 & 19 & 17 & 17 & 17 & 18 & 19 \\
\cmidrule{2-12}
& Rearrange Coke & 36 & 21 & 21 & 20 & 19 & 22 & 21 & 25 & 22 & 26 \\
\midrule
\textbf{RoboBaton MINI} & Pick Cup & 17 & 15 & 16 & 14 & 13 & 15 & 15 & 14 & 16 & 17 \\
\cmidrule{2-12}
& Open Container & 10 & 11 & 10 & 11 & 11 & 12 & 11 & 12 & 12 & 12 \\
\bottomrule
\end{tabular}
\end{table*}

\begin{figure}[htbp]
    \centering
    \includegraphics[width=.7\linewidth]{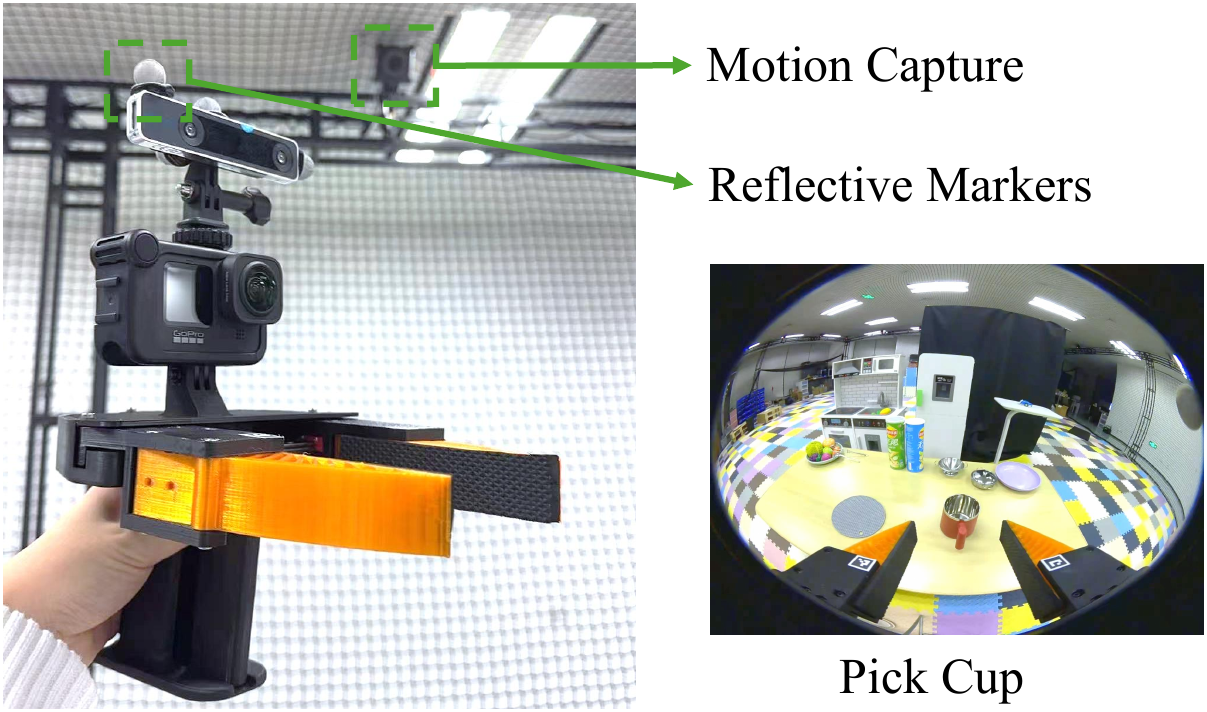}
    \caption{\textbf{Left}: Four reflective markers attached to the FastUMI handheld device, tracked by an optical motion-capture system for ground-truth trajectory collection. \textbf{Right}: Example scenarios from the ``Pick Cup'' task.}
    \label{fig:capture}
\end{figure}

In TABLE~\ref{tab:trajectory_errors}, we summarize the pose estimation errors of both T265 and MINI across three representative tasks: ``Pick Cup,'' a straightforward pick-and-place action (as shown in Fig.~\ref{fig:capture} Right); ``Open container,'' which involves hinged motion and partial occlusion; and ``Rearrange Coke,'' a more complex scenario with substantial occlusion. 
To establish ground-truth trajectories, four reflective markers are affixed to the handheld device and tracked by an optical motion-capture system (Fig.~\ref{fig:capture} Left).
Simultaneously, the device poses are recorded through our data collection pipeline. 
All data streams are synchronized within ROS via unified timestamps, and ten trajectories per sensor are collected for each task. 
The \textit{evo} toolkit is used to compute all reported errors~\cite{grupp2017evo}.

\begin{table*}[t]
\centering
\caption{Success rates for DP and ACT in different tasks, sorted by manipulation type.}\label{tab:baseline}
\begin{adjustbox}{max width=\textwidth}
\begin{tabular}{@{}c l c c c@{}}
\toprule
\textbf{Index} & \textbf{Task} & \textbf{Manipulation Type} &  \makecell{\textbf{Success Rate (\%) of DP}\\ \textbf{(Relative TCP)}} & \makecell{\textbf{Success Rate (\%) of ACT}\\ \textbf{(Absolute Joint)}} \\
\midrule
\midrule
1 & Open Container   & Hinged      & 93.33 & 86.67 \\
2 & Open Roaster     & Hinged      & 80.00 & 86.67 \\
3 & Open Drawer      & Hinged      & 53.33 & 80.00 \\
4 & Open Suitcase    & Hinged      & 40.00 & 86.67 \\
\midrule
5 & Rearrange Coke   & Pick-Place  & 80.00 & 86.67 \\
6 & Fold Towel       & Pick-Place  & 93.33 & 73.33 \\
7 & Pick Bear        & Pick-Place  & 80.00 & 20.00 \\
8 & Unplug Charger   & Pick-Place  & 86.67 & 86.67 \\
9 & Pick Lid         & Pick-Place  & 53.33 & 93.33 \\
10 & Pick Pen         & Pick-Place  & 53.33 & 20.00 \\
\midrule
11 & Sweep Trash      & Pick-Push   & 46.67 &  6.67 \\
\midrule
12 & Open Ricecooker  & Button Press & 20.00 & 80.00 \\
\bottomrule
\end{tabular}
\end{adjustbox}
\end{table*}

\begin{figure}[htbp]
    \centering
    \includegraphics[width=.8\linewidth]{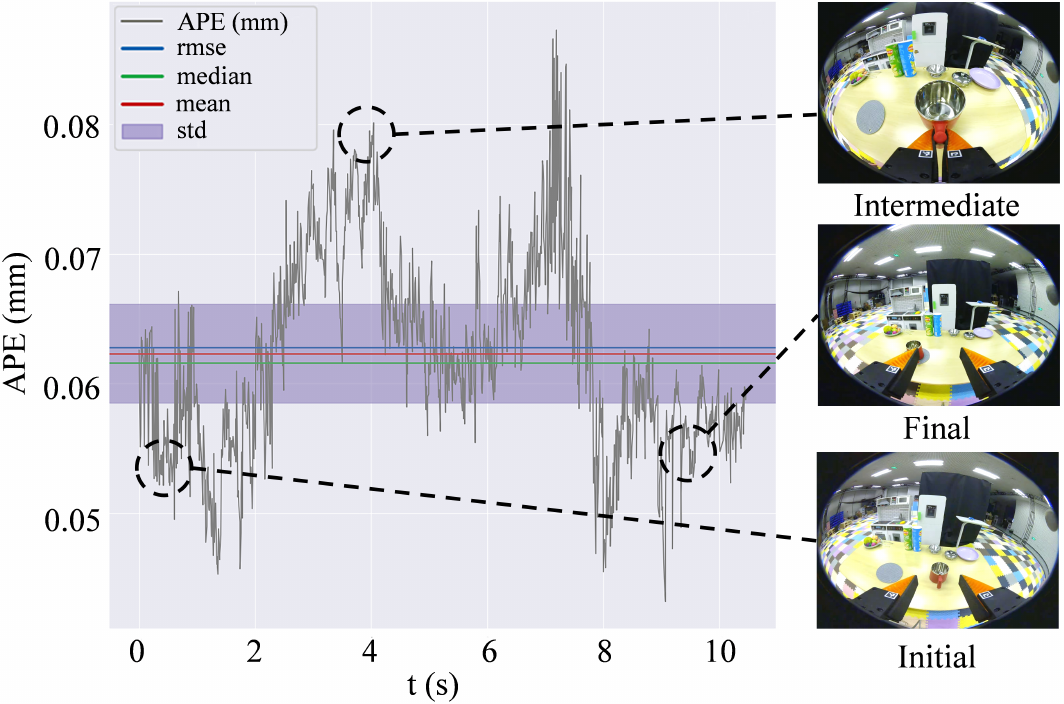}
    \caption{Representative T265 VIO error over time during the “Pick Cup” task. 
    Error peaks appear when the gripper nears the table and occludes visual features, then recover once it returns to the original viewpoint.}
    \label{fig:t265}
\end{figure}

In the ``Pick Cup'' scenario, where occlusion is minimal, T265 achieves an average positioning error of 10.5 mm, while MINI's error averages 15.2 mm. 
In the ``Open Container,'' T265's error increases to 17.7 mm, reflecting the partial obstruction of its field of view, whereas MINI's error decreases to 11.2 mm. 
T265's performance degrades further in the ``Rearrange Coke,'' where placing an object inside a cabinet induces significant occlusion.
These findings indicate that T265 is particularly susceptible to severe visual obstruction at close range. 
%, especially under rapid accelerations and frequent orientation changes; prolonged sequences further amplify drift, reducing precision toward the end of each trajectory
In contrast, MINI demonstrates relatively stable cross-task performance\textemdash albeit with slightly reduced accuracy in low-occlusion scenarios.
T265 offers superior localization when visual inputs are largely unobstructed, whereas MINI exhibits more consistent performance under varying levels of occlusion.
We also observe that the VIO error typically remains low at the beginning and end of each trajectory but grows noticeably in the middle.
Fig.~\ref{fig:t265} illustrates this pattern for T265 during the ``Pick Cup'' task: as the gripper moves closer to the table, occlusion reduces visible features and causes two pronounced error peaks. 
During intermediate movement, partial visibility leads to moderate errors, though still higher than at the outset. 
By the final stage, returning to the original viewpoint restores abundant features, and loop-closure mechanisms recover tracking accuracy to near-initial levels.

\subsection{Baseline Performance}\label{sec:baseline}

We conduct a comparative study of two baseline approaches for policy inference\textemdash ACT with absolute joint-space outputs and DP with relative TCP-based outputs\textemdash across 12 diverse manipulation tasks.
Each task is trained on \textbf{200 demonstrations}, randomly selected from our open-source dataset.
Among the 200 pieces of data, every 50 pieces are a group. The robotic arm's initial configuration and environment in one group are fixed, while the target object's position varies within a predefined range.
Across groups, both the arm's starting pose and the scene arrangement are altered.
Each algorithm is trained on the same training data.
During testing, we use object poses that appear in the training set but \textbf{place them in new scene contexts}.
Each task is attempted \textbf{15 times}, and success rates are recorded.

The results are shown in TABLE~\ref{tab:baseline}.
Both ACT and DP achieve relatively high success rates on most tasks, indicating that the collected dataset is sufficiently diverse and general to support different policy representations.
Notably, tasks involving substantial occlusion (e.g., ``Rearrange Coke'' and ``Open Container'') can still be effectively handled, suggesting that our data collection strategy is robust to partial visibility. 

In tasks requiring precise depth estimation \textemdash such as ``Open Drawer,'' ``Pick Lid,'' and ``Open Ricecooker''\textemdash the baseline \textbf{DP} algorithm's limitations become evident.
In particular, DP struggled with pressing actions in ``Open Ricecooker,'' where small deviations in relative motion can prevent successful button presses.
By contrast, in the baseline \textbf{ACT} algorithm, tasks like ``Open Suitcase,'' exhibit more accurate depth reasoning but less sensitivity to specific trajectory requirements.
In ``Pick Bear,'' ACT occasionally generates joint configurations unseen during training (e.g., producing a fully inverted gripper posture when the dataset predominantly showed a downward TCP orientation),  which highlights a known limitation of the original ACT approach, \emph{typically reliant on third-person viewpoints for global state estimation.}
Similar issues arise in ``Pick Pen'' and ``Sweep Trash,'' with the latter also revealing a workspace mismatch: ACT's absolute joint predictions sometimes yield unreachable targets if training data contained trajectories that exceeded the xArm's operational envelope.
By contrast, DP's incremental relative-position strategy partly alleviated this problem, although multi-step tasks like ``Sweep Trash'' remain challenging for both models.

\subsection{Algorithmic Enhancements}\label{sec:enhancement}
We evaluate our algorithmic refinements on two tasks\textemdash ``Pick Lid'' and ``Open Ricecooker''\textemdash where the baseline DP approach most struggles with depth estimation.
We retain the same training data and parameters as in the previous experiment for both ACT and DP, and employ the same testing protocol.

In the \textbf{Depth-Enhanced DP}, we incorporate depth information into the original DP algorithm.
As shown in TABLE~\ref{tab:dp_vs_dp_depth},  success rates  increase by 26.67\% on ``Pick Lid'' and 73.33\% on ``Open Ricecooker.'' 
These results highlight the importance of depth cues for precise object manipulation tasks, particularly those requiring accurate vertical alignment or force application.
For ACT, we introduce two variants called \textbf{Smooth-ACT} and \textbf{PoseACT}, which incorporate GRU-based temporal modeling and integrates TCP (end-effector state) inputs.
We demonstrate that these two variants yield substantial improvements in success rate compared to the original ACT, as shown in TABLE~\ref{tab:task_success_comparison}. 
Furthermore, we explore relative TCP to reduce dependence on absolute coordinates, aiming to capture the shape and dynamics of the trajectory more robustly. 
As the table indicates, this enhancement performs well on tasks featuring extended or repetitive trajectories (e.g., ``Sweep Trash''). 
However, for tasks requiring precise height estimation (such as ``Pick Bear''), removing absolute pose information can degrade vertical positioning accuracy, underscoring a trade-off between relative and absolute coordinate representations.

\begin{table}[htbp]
    \centering
    \caption{Comparison of success rates for DP and DP + Depth in tasks with significant depth-estimation challenges.}
    \label{tab:dp_vs_dp_depth}
    \begin{adjustbox}{max width=\textwidth}
    \begin{tabular}{@{}lcc@{}}
    \toprule
    \textbf{Task} & \makecell{\textbf{Success Rate (\%) of} \\ \textbf{Original DP}} & \makecell{\textbf{Success Rate (\%) of} \\ \textbf{Depth-Enhanced DP}} \\
    \midrule
    Pick Lid & 53.33\% & 80.00\% \\
    Open Ricecooker & 20.00\% & 93.33\% \\
    \bottomrule
    \end{tabular}
    \end{adjustbox}
\end{table}

\begin{table}[htbp]
    \centering
    \caption{Comparison of success rates for different ACT variants across representative tasks.}
    \label{tab:task_success_comparison}
    \begin{adjustbox}{max width=\linewidth}
    \begin{tabular}{@{}l cc cc@{}}
    \toprule
    & \multicolumn{2}{c}{\textbf{Joint}} & \multicolumn{2}{c}{\textbf{TCP}} \\
    \cmidrule(lr){2-3} \cmidrule(lr){4-5}
    \textbf{Task} & \textbf{ACT} & \textbf{Smooth-ACT} & \makecell{\textbf{PoseACT}\\ \textbf{(Absolute)}} & \makecell{\textbf{PoseACT}\\ \textbf{(Relative)}} \\
    \midrule
    Pick Bear    & 20.00\% & 60.00\% & 80.00\% & 73.33\% \\
    Sweep Trash  &  6.67\% & 26.67\% & 53.33\% & 60.00\% \\
    \bottomrule
    \end{tabular}
    \end{adjustbox}
\end{table}

% \begin{table}[htbp]
%     \centering
%     \caption{Comparison of success rates for different ACT variants across representative tasks.}
%     \label{tab:task_success_comparison}
%     \begin{adjustbox}{max width=\textwidth}
%     \begin{tabular}{@{}lcccc@{}}
%     \toprule
%     \textbf{Task} & \textbf{ACT + Joint} & \textbf{ACT + Joint + GRU} & \textbf{ACT + Absolute TCP + GRU} & \textbf{ACT + Relative TCP + GRU} \\
%     \midrule
%     Pick Bear & 20.00\% & 60.00\% & 80.00\% & 73.33\% \\
%     Sweep Trash & 6.67\% & 26.67\% & 53.33\% & 60.00\% \\
%     \bottomrule
%     \end{tabular}
%     \end{adjustbox}
% \end{table}

\subsection{Additional Factors}\label{sec:additional}

We further investigate the influence of camera configurations and training data size on policy inference performance. TABLE~\ref{tab:camera_performance} compares different camera setups for both pick-and-place and hinged operations. 
For each configuration (i.e., camera model and lens type), we collected 50 demonstrations under identical scene settings, with only the target object's position randomly varied within a small range. 
All trajectories were obtained via direct teleoperation. 
The original ACT algorithm is then evaluated on object positions seen during training but under new trials, each repeated 15 times to compute the success rate.
Notably, \emph{a fisheye lens at the end-effector achieves performance comparable to multi-view setups}, potentially because its wide field of view captures richer contextual information for decision-making.

Next, to assess how the amount of training data affects generalization, we conducted an experiment on a ``Pick Cup'' task with 200, 400, and 800 demonstrations (TABLE~\ref{tab:pick_cup_success_rates}). 
In this scenario, the cup and coaster were each placed in five distinct positions, repeated three times with different handle orientations. 
The original ACT model must learn not only positional information but also handle orientation to generate an appropriate grasp trajectory. 
As the dataset grew larger, success rates significantly improved, indicating that data abundance bolsters the model's capacity to generalize across varied object placements and orientations.

% 鱼眼相机画幅成功率对比
\begin{table}[htbp]  % 使用table而非table*，使其成为单列浮动体
    \centering
    \caption{Comparison of task performance under varying camera setups (lens type and viewpoint).}
    \label{tab:camera_performance}
    \begin{adjustbox}{max width=\linewidth}  % 控制宽度适应单栏
    \begin{tabular}{l cc}
    \toprule
    %=========================%
    % 第一行相机配置
    %=========================%
    & \textbf{D435i} & \textbf{GoPro with Flat Lens} \\
    & (First-Person) & (First-Person) \\
    \midrule
    Pick Bear       & 0\%    & 6.67\%  \\
    Open Container  & 0\%    & 93.33\% \\
    \midrule
    %=========================%
    % 第二行相机配置
    %=========================%
    & \textbf{D435i} & \textbf{GoPro with Fisheye Lens} \\
    & (First-Person\&Third-person) & (First-Person) \\
    \midrule
    Pick Bear       & 86.67\% & 80.00\% \\
    Open Container  & 100.00\% & 100.00\% \\
    \bottomrule
    \end{tabular}
    \end{adjustbox}
\end{table}

\begin{table}[htbp]
    \centering
    \caption{Success rates in the ``Pick Cup'' task using different training dataset sizes.}
    \label{tab:pick_cup_success_rates}
    \begin{adjustbox}{max width=\textwidth}
    \begin{tabular}{@{}lcccc@{}}
    \toprule
    \textbf{Task} & \textbf{Data Size (200)} & \textbf{Data Size (400)} & \textbf{Data Size (800)} \\
    \midrule
    Pick Cup & 20.00\% & 26.67\% & 53.33\% \\
    \bottomrule
    \end{tabular}
    \end{adjustbox}
\end{table}

\section{Limitations}
While FastUMI demonstrates effective policy execution across diverse tasks, several limitations remain:

\vspace{0.5em} \noindent\textbf{1) Limited Sensing Modalities}. FastUMI currently relies on visual data, which may prove insufficient for tasks requiring precise force or tactile feedback\textemdash such as handling fragile objects.
Integrating tactile or force sensors could enable richer environmental representations and more robust policy learning, particularly for tasks necessitating delicate or high-precision interactions.

\vspace{0.5em} \noindent\textbf{2) Restricted Robot Compatibility}. Although FastUMI accommodates single-arm or dual-arm platforms, it is not yet adapted for more complex morphologies, including mobile manipulators requiring whole-body control. 
Future endeavors could focus on expanding the hardware and software ecosystem to support advanced platforms with larger workspaces and non-static bases.

\vspace{0.5em} \noindent\textbf{3) Wired Data Transfer}. Reliance on wired connections constrains portability and limits field applications where mobility or standalone operation is pivotal. 
A wireless solution with onboard processing or seamless network connectivity would greatly expand FastUMI's applicability and facilitate broader deployment.

% \vspace{0.5em} \noindent\textbf{4) Data Quality Assessment}. While we have introduced basic data consistency checks (e.g., trajectory validation), these alone may not ensure the overall quality required for reliable policy training. 
% More comprehensive verification—such as automated anomaly detection or real-time feedback—would help maintain dataset integrity and enhance training outcomes.

\section{Related Work}
\subsection{Data Collection Methods}
High-quality data is fundamental to the success of learning algorithms~\cite{belkhale_data_2023}. 
% How to efficiently collect these high-quality data becomes an important area for researchers to explore naturally. 
Here, we introduce several data collection systems and compare them to our Fast-UMI.
\textbf{Teleoperated systems} represent one of the most widely adopted methodologies for data collection in imitation learning. 
This approach enables researchers to intuitively gather demonstration data, establishing a direct correspondence between observed visual inputs and associated actions ~\cite{zhang2018deepimitationlearningcomplex}. 
Various control interfaces, including  AR controllers~\cite{park_dexhub_2024,chen2024arcap}, haptic controllers~\cite{owan_faster_2020,ding_bunny-visionpro_2024}, 3D spacemouses~\cite{liu_robot_2024}, and newly explored leader-follower systems~\cite{wu2024gello} are developed to build teleoperated systems.
However, these systems inherently depend on real robotic arms during data collection. 
Additionally, hardware-specific constraints often necessitate modifications to enable cross-platform compatibility, significantly reducing efficiency. 
In contrast, Fast-UMI requires only a handheld device, enabling portable and flexible data collection.
% Also, Fast-UMI records the TCP poses as proprioception information instead of joint angles, which allows the data to be reused on any robot platform.

An alternative data collection paradigm involves capturing multi-view \textbf{human demonstration videos}. 
Robots can extract actionable knowledge from these recordings by leveraging adversarial learning objectives~\cite{mees_adversarial_2020}, contextualized annotations~\cite{sontakke_roboclip_2023}, and hybrid CNN-probabilistic parsing techniques~\cite{yang_robot_2015}. 
This approach circumvents the need for physical robotic platforms and facilitates the construction of reusable datasets. 
However, it presents several inherent limitations. 
Since the action data are inferred from raw videos, these actions sometimes may not precisely reflect the true actions, which hinders the formation of generalizable policies. 
Furthermore, the embodiment mismatch remains a persistent challenge, as discrepancies between the domain in which data is collected and the deployment environment can lead to policy failures~\cite{eze_learning_2024}. 
In contrast, Fast-UMI directly collects precise action information during demonstrations and minimizes domain shift by aligning video observation of wrist-mounted cameras on both the hand-held device and the on-robot device.

\textbf{Sensor-enhanced interfaces} (i.e., handheld grippers) offer a promising alternative for data collection, addressing some of the aforementioned challenges. 
However, obtaining precise TCP pose information remains nontrivial. 
Existing solutions incorporate SLAM-based estimation from video streams~\cite{chi2024universal}, motion capture systems~\cite{wang2023mimicplaylonghorizonimitationlearning}, and vision-based tracking algorithms~\cite{song_grasping_2020}.
These techniques, however, often necessitate extensive post-processing or rely on fixed infrastructure, reducing overall efficiency.
In contrast, Fast-UMI employs the T265 to directly capture accurate pose data, eliminating the need for cumbersome SLAM pipelines or motion capture systems.
Additionally, its wrist-mounted gopro camera records high-resolution visual data at variable frame rates, providing a rich observational dataset to support policy learning.

\subsection{Imitation Learning}
Unlike methods that heavily rely on human programming~\cite{aeronautiques1998pddl} and task-specific reward functions~\cite{arulkumaran2017deep, daniel2014active}, Imitation Learning (IL) enables robots to autonomously perform tasks by learning from expert demonstrations~\cite{schaal1999imitation,hussein2017imitation,fang2019survey,stepputtis2020language,zare2024survey}. 
With the large-scale collection of robotic manipulation datasets in recent years~\cite{o2024open,walke2023bridgedata,brohan2022rt,khazatsky2024droid,kalashnikov2018scalable,fang2024rh20t}, IL has been widely adopted in robotic manipulation, demonstrating remarkable performance across diverse task domains. 
Depending on the nature of the collected data, IL algorithms can leverage real-robot demonstrations~\cite{zhao2023learning,fu2024mobile,shi2024yell},   video-based observations without explicit action labels~\cite{wang2023mimicplaylonghorizonimitationlearning,bahl2022humantorobotimitationwild}, or data obtained from decoupled handheld tracking devices~\cite{chi2024universal,doshi2023hand,sanches2023scalable}. 
Furthermore, in dexterous hand manipulation tasks, IL has been extended to learn from human hand motion demonstrations~\cite{wang2024dexcap,chen2024arcap}. 
The ACT algorithm applies imitation learning to absolute joint pose data collected from robotic arms and utilizes temporal ensemble techniques over fixed-length action sequences to enable smooth and autonomous dual-arm control~\cite{zhao2023learning}.
The work~\cite{shi2024yell} integrates a language modality into the ACT algorithm, fine-tuning a vision-language model to facilitate language-based interaction with dual-arm robots.  
DP generates actions in the robotic action space through a conditional denoising process, offering advantages such as expressing multimodal action distributions, handling high-dimensional output spaces, and providing stable training~\cite{chi2023diffusion}. 
UMI demonstrates that UMI-like data can be effectively used to train diffusion policy, and yield promising results~\cite{chi2024universal}. 
% In our work, we validate our system's performance using data collected through ACT~\cite{zhao2023learning} and Diffusion Policy~\cite{chi2023diffusion} and explore the impact of different data types and the inclusion of UMI-like depth information on these algorithms. 
In our work, we validate our system's performance using data collected through ACT~\cite{zhao2023learning} and DP~\cite{chi2023diffusion}. 
We further analyze the characteristics of the data collected by Fast-UMI and evaluate its impact on these algorithms. 
Based on this analysis, we implement a series of optimizations and adaptations, enhancing the performance of these algorithms when applied to Fast-UMI data.
% This research will contribute to the adaptation of future algorithms to UMI-like data, further expanding the development of the UMI-like dataset community. 

\section{Conclusion}
In this work, we introduce \emph{FastUMI}, a redesigned system built on the original UMI to streamline real-world data collection for robotic manipulation. 
Our hardware modifications enable quick deployment across diverse arms and grippers, removing dependencies on specialized components. 
By replacing complex SLAM with T265-based tracking, FastUMI reduces calibration overhead and maintains robust performance despite occlusions. 
We also open-source a dataset of 10,000 real-world demonstrations spanning 22 everyday tasks. 
Experiments confirm that FastUMI lowers costs, simplifies deployment, and supports large-scale data-driven policy learning. 
Future work will focus on integrating richer sensing modalities and extending FastUMI to more complex platforms.

% \section*{Acknowledgments}

%% Use plainnat to work nicely with natbib. 

\bibliographystyle{plainnat}
\bibliography{reference}

\newpage
\section{Appendix}~\label{sec:appendix}

\subsection{RealSense T265 vs. RoboBaton MINI}\label{appendix_t265_mini}
We compare the main parameters between T265 and MINI.

\begin{table}[htbp]
\vspace{-1.0em}
\renewcommand{\arraystretch}{1.3}
\caption{Device Specifications Comparison}
\centering
\label{tab:spec_comparison}
\begin{tabular}{@{}lcc@{}}
\hline\hline \\[-3mm]
 & \textbf{T265} & \textbf{MINI} \\ \\[-3mm] \hline
Output Frequency (Hz) & 200 & 20 \\ \hline
Accuracy (mm) & 10 & 10 \\ \hline
FOV & 163°(D) & \begin{tabular}[c]{@{}c@{}}164.7°(D)\\ 164.7°(H)\\ 123.8°(V)\end{tabular} \\ \hline
Resolution & 848×800 & 640×480 \\ \hline
Weight (g) & 55 & 68 \\ \hline
Dimensions (mm) & 108×24.5×12.5 & 101.6×32.25×17.70 \\ \hline
SDK & \begin{tabular}[c]{@{}c@{}}Windows/Linux\\ ROS1\end{tabular} & \begin{tabular}[c]{@{}c@{}}Windows/Linux\\ HTTP/ROS2\end{tabular} \\ 
[1mm] \hline\hline
\end{tabular}
\vspace{-0.5em}
\end{table}

\begin{figure}[!h]
    \centering
    \includegraphics[width=0.55\linewidth]{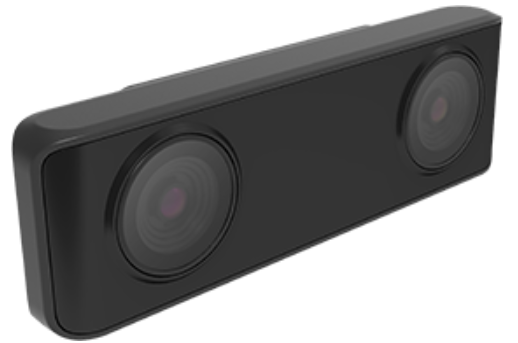}
    % \vspace{-0.5em}
    \caption{The RoboBaton MINI product image.}
    \label{fig:mini}
\end{figure}

\subsection{FastUMI Dataset}\label{appendix_dataset}
Our dataset is composed of more than 10000 demonstrations from 22 daily tasks. The dataset has been split into smaller parts. 
Users need to merge the files after downloading to reconstruct the original dataset. Each file is named with its corresponding task name and contains no more than 50 HDF5 files. 
Each HDF5 file corresponds to a single episode and encapsulates both observational data and actions. 
Below is the hierarchical structure of the HDF5 file:
\begin{verbatim}
episode_<idx>.hdf5
|-- observations/
|   |-- images/
|   |   `-- <camera_name_1> (Dataset)
|   `-- qpos (Dataset)
|-- action (Dataset)
`-- attributes/
    `-- sim = False
\end{verbatim}
The variable ``sim'' indicates whether the data was recorded in simulation (True) or real-world (False). 
The ``images'' stores image data from cameras as uint8 and has a shape of (num\_frames, height=1920, width=1080, channels=3). 
The ``qpos'' stores position and orientation data for each timestep and has a shape of (num\_timesteps, 7), where the 7 columns correspond to [Pos X, Pos Y, Pos Z, Q\_X, Q\_Y, Q\_Z, Q\_W]. 
The ``actions'' stores action data corresponding to each timestep. 
In this script, actions mirror the qpos data.

\end{document}